\documentclass[pre,floatfix,superscriptaddress,twocolumn,9pt]{revtex4-1}

\usepackage{amsmath,amssymb}
\usepackage{mathrsfs}
\usepackage{mathtools}
\usepackage{graphicx}
\usepackage{color}
\usepackage{dcolumn}
\usepackage[normalem]{ulem}
\usepackage{algorithm}
\usepackage{subfig}
\usepackage{algpseudocode}
\usepackage{comment}

\algdef{SE}[DOWHILE]{Do}{doWhile}{\algorithmicdo}[1]{\algorithmicwhile\ #1}

\newcommand{\p}[1]{\left(#1\right)}
\newcommand{\bk}[2]{\langle #1 \mid #2\rangle} 

\newcommand{\abs}[1]{\left| #1 \right|} 
\newcommand{\pd}[2]{\frac{\partial #1}{\partial #2}} 
 
\newcommand{\Expect}[1]{\mathrm{E}\left[#1\right]}

\usepackage[disable]{todonotes} 				
\newcommand*{\rood}[1]{#1}
 \newcommand*{\blauw}[1]{#1}

\newcommand{\figurewidth}{.45\textwidth}

\begin{document}
	
	\title{Nonlinear Discovery of Slow Molecular Modes using State-Free Reversible VAMPnets}
	
	\author{Wei Chen}
	\affiliation{
		Department of Physics, %
		University of Illinois at Urbana-Champaign, %
		1110 West Green Street, Urbana, Illinois 61801
	}
	
	\author{Hythem Sidky}
	\affiliation{%
		Institute for Molecular Engineering, %
		5640 South Ellis Avenue, %
		University of Chicago, %
		Chicago, Illinois 60637%
	}
	
	\author{Andrew L. Ferguson}
	\email{Author to whom correspondence should be addressed: \mbox{andrewferguson@uchicago.edu}}
	\affiliation{%
		Institute for Molecular Engineering, %
		5640 South Ellis Avenue, %
		University of Chicago, %
		Chicago, Illinois 60637%
	}
	
	\begin{abstract}
		\noindent The success of enhanced sampling molecular simulations that accelerate along collective variables (CVs) is predicated on the
		availability of variables coincident with the slow collective motions governing the long-time conformational dynamics of 
		a system. It is challenging to intuit these slow CVs for all but the simplest molecular systems, and their data-driven discovery
		directly from molecular simulation trajectories has been a central focus of the molecular simulation community to both unveil
		the important physical mechanisms and to drive enhanced sampling. In this work, we introduce state-free reversible VAMPnets (SRV)
		as a deep learning architecture that learns nonlinear CV approximants to the leading slow eigenfunctions of the spectral decomposition of the transfer operator that evolves equilibrium-scaled probability distributions through time. Orthogonality of the learned CVs is naturally imposed within network training without added
		regularization. The CVs are inherently explicit and differentiable functions of the input coordinates making them well-suited to use in enhanced sampling calculations. We demonstrate the utility of SRVs in capturing 
		parsimonious nonlinear representations of complex system dynamics in applications to 1D and 2D toy systems where the true eigenfunctions are exactly calculable and to molecular dynamics simulations of alanine dipeptide and the WW domain protein. 
	\end{abstract}
	
	\maketitle

	\section{Introduction} \label{sec:intro}
	
	Molecular dynamics (MD) simulations have long been an important tool for studying molecular systems by 
	providing atomistic insight into physicochemical processes that cannot be easily obtained through experimentation.  
	A key step in extracting kinetic information from molecular simulation is the recovery of the slow dynamical modes 
	that govern the long-time evolution of system coordinates within a low-dimensional latent space. The variational approach to conformational
	dynamics (VAC)~\cite{noe2013variational, nuske2014variational} has been successful in providing a mathematical framework 
	through which the eigenfunctions of the underlying transfer operator can be estimated \cite{Noe2017, klus2018data}. 
	
	A special case 
	of VAC which estimates linearly optimal slow modes from mean-free input coordinates is known as time-lagged independent
	component analysis (TICA) \cite{perez2013identification, noe2013variational, nuske2014variational, noe2015kinetic, noe2016commute,
		perez2016hierarchical, schwantes2013improvements, klus2018data}. TICA is a widely-used approach that has become a standard step in the Markov state modeling pipeline~\cite{husic2018markov}. However, it is restricted to form linear combinations of the input coordinates and is unable to learn nonlinear transformations that are typically required to recover high resolution kinetic models of all but the simplest molecular systems. Schwantes \textit{et al.}\ address this limitation by applying the
	kernel trick with TICA to learn non-linear functions of the input coordinates \cite{schwantes2015modeling}. A special case of a radial basis function kernels was realized by No\'e and Nuske in the direct application of VAC using Gaussian functions \cite{noe2013variational}. Kernel TICA (kTICA), however, suffers from a number of drawbacks that have precluded its broad adoption. First, its implementation requires $O(N^2)$ memory usage and $O(N^3)$ computation time for a dataset of size $N$, which becomes intractable for large datasets. In practice, this issue can be adequately alleviated by selecting a small number of landmark points to which kernel TICA is applied, and then constructing interpolative projections of the remaining points using the Nystr\"om extension~\cite{harrigan2017landmark}. Second, kTICA is typically highly sensitive to the kernel hyperparameters~\cite{harrigan2017landmark}, and extensive hyperparameter tuning is typically required to obtain acceptable results. Third, the use of the kernel trick and Nystr\"om extension compromises differentiability of the latent space projection. Exact computation of the gradient of any new point
	 requires the expensive calculation of kernel function $K(x,y)$ of new data $x$ with respect to all points $y$ in the training set, and gradient estimations based only on the landmarks are inherently approximate~\cite{m2017tica}. Due to their high cost and/or instability of these gradient estimates, the slow modes estimated through kTICA are impractical for use as collective variables (CVs) for enhanced sampling or reduced-dimensional modeling approaches that require exact gradients of the CVs with respect to the input coordinates.
	
	Deep learning offers an attractive alternative to kTICA as means to solve these challenges. Artificial neural networks are capable of learning nonlinear functions of arbitrary complexity~\cite{hassoun1995fundamentals,chen1995universal}, are generically scalable to large datasets with training scaling linearly with the size of the training set, the network predictions are relatively robust to the choice of network architecture and activation functions, and exact expressions for the derivatives of the learned CVs with respect to the input coordinates are available by automatic differentiation~\cite{bastien2012theano, baydin2018automatic,paszke2017automatic, schmidhuber2015deep}. A number of approaches utilizing artificial 
	neural networks to approximate eigenfunctions of the dynamical operator have been proposed. Time-lagged autoencoders~\cite{wehmeyer2018time} utilize auto-associative neural networks to reconstruct a time-lagged signal, with 
	suitable collective variables extracted from the bottleneck layer. Variational dynamics
	encoders (VDEs)~\cite{hernandez2018variational} combine time-lagged reconstruction loss and autocorrelation maximization within 
	a variational autoencoder. While the exact relationship between the regression approach employed in time-lagged autoencoders and the VAC framework 
	is not yet formalized \cite{wehmeyer2018time}, variational autoencoders (VAEs) have already been studied as estimators of 
	the dynamical propagator~\cite{hernandez2018variational} and in furnishing collective variables for enhanced
	sampling~\cite{sultan2018transferable}. The most pressing limitation of VAE approaches to date is their restriction to the estimation of the single leading
	eigenfunctions of the dynamical propagator. The absence of the full spectrum of slow modes fails to expose the full richness of the underlying dynamics, and limits enhanced sampling calculations in the learned CVs to acceleration along a single coordinate that may be insufficient to drive all relevant conformational interconversions. 
	
	In this work, we propose a deep-learning based method to estimate the slow dynamical modes that we term \rood{state-free reversible VAMPnets (SRVs)}. SRVs take advantage of the VAC framework using a neural network architecture and loss function to recover the leading modes of the spectral hierarchy of eigenfunctions of the transfer operator that evolves equilibrium-scaled probability distributions through time. In a nutshell, the SRV discovers a nonlinear featurization of the input basis to pass to the VAC framework for estimation of the leading eigenfunctions of the transfer operator. 
	
	\rood{This approach shares much technical similarity with, and was in large part inspired by, the elegant VAMPnets approach developed by No\'e and coworkers \cite{mardt2018vampnets} and deep canonical correlation analysis (DCCA) approach developed by Livescu and coworkers \cite{andrew2013deep}. Both approaches employ Siamese neural networks to discover nonlinear featurizations of an input basis that are optimized using a VAMP score. VAMPnets differ from DCCA in optimizing a VAMP-2 rather than VAMP-1 score, making it better suited to applications to time series data due to its theoretical grounding in the Koopman approximation error \cite{mardt2018vampnets}. VAMPnets seek to replace the entire MSM construction pipeline of featurization, dimensionality reduction, clustering, and construction of a kinetic model. The objective of SRVs is not to perform direct state space partitioning but rather to learn continuous nonlinear functions of the input data to generate a nonlinear basis set with which to approximate the eigenfunctions of the transfer operator. The design of SRVs differs from that of VAMPnets in two important ways to support this goal. Indeed the name given to our approach is intended to both indicate its heritage with VAMPnets and these distinguishing features. First, the SRV optimizes the VAC as a variational principle for stationary and reversible processes \cite{noe2013variational, nuske2014variational}, whereas VAMPnets employ the more general VAMP principle that applies to non-stationary and non-reversible processes \cite{mardt2018vampnets}. As such, SRVs are designed for applications to molecular systems obeying detailed balance where the VAC permits us to take molecular trajectories that may not strictly obey detailed balance and make a biased estimation of the slow eigenfunctions of  the reversible dynamics rather than a less biased estimator of the possibly non-reversible dynamics contained in the finite data. Second, VAMPnets employ softmax activations within their output layers to generate $k$-dimensional output vectors that can be interpreted as probability assignment to each of $k$ metastable states. This network architecture achieves nonlinear featurization of the input basis and soft clustering into metastable states. The output vectors are subsequently optimized using the VAMP principle to furnish a kinetic model over these soft/fuzzy states by approximating the eigenfunctions of the transfer operator over the states. Importantly, even though the primary objective of VAMPnets is clustering, the soft state assignments can be used to approximate the transfer operator eigenfunctions using a reweighting procedure. However, since the soft state assignments produced by softmax activations of the output layer are constrained to sum to unity, there is a linear dependence that requires $(k+1)$ output components to identify the $k$ leading eigenfunctions. The second distinguishing feature of SRVs is thus to employ linear or nonlinear activation functions in the output layer of the network. By eschewing any clustering, the SRV is better suited to directly approximating the transfer operator eigenfunctions as its primary objective, although clustering can also be performed in the space of these eigenfunctions in a post-processing step. This seemingly small change has a large impact in the success rate of the network in successfully recovering the transfer operator eigenfunctions. Training  neural networks is inherently stochastic and it is standard practice to train multiple networks with different initial network parameters and select the best. Numerical experiments on the small biomolecule alanine dipeptide (see Section III.C) in which we trained 100 VAMPnets and 100 SRVs for 100 epochs employing optimal learning rates showed that both VAMPnets and SRVs were able to accurately recover the leading eigenfunctions and eigenvalues in quantitative agreement with one another, but that VAMPnets exhibited a 29\% success rate in doing so compared to 70\% for SRVs. Accordingly, the architecture of SRVs is better suited to direct estimation of the transfer operator eigenfunctions and may be preferred when the objective is to estimate these functions as continuous, explicit, and differentiable functions of the input coordinates that can be used to infer the mechanisms of molecular conformational transitions, employed directly in enhanced sampling calculations, and passed to standard MSM construction pipelines to perform microstate clustering and estimation of discrete kinetic models.}
	
	The structure of this paper is as follows. We first derive the theoretical foundations of the SRV as a special case of VAC, and then demonstrate its efficacy against kTICA and state-of-the-art TICA-based MSMs in applications to 1D and 2D toy systems where the true eigenfunctions are known and in molecular simulations of alanine dipeptide and WW domain.

	\section{Methods} \label{sec:method}
	
	We first recapitulate transfer operator theory and the variational approach to conformational dynamics (VAC)~\cite{noe2013variational, nuske2014variational, schutte2001transfer, prinz2011markov, schwantes2015modeling}, choosing to borrow the notational convention from Ref.~\cite{prinz2011markov}. We then demonstrate how the VAC specializes to TICA, kTICA, and SRVs.
	
\subsection{Transfer operator theory}\label{th:transfer_op}
	Given the probability distribution of a system configuration $p_t(x)$ at time $t$ and the equilibrium probability distribution
	$\pi(x)$, we define $u_t(x)=p_t(x)/\pi(x)$ and the transfer operator $\mathscr{T}_t=\mathscr{T}_t(\tau)$, known formally as the Perron-Frobenius operator or propagator with respect to the equilibrium density \cite{klus2018data}, such that 
	\begin{equation}\label{transfer}
	u_{t+\tau}(x)=\mathscr{T}_t\circ u_t(x)=\frac{1}{\pi(x)} \int dy\ p^t_\tau(y,x)u_t(y)\pi(y)
	\end{equation}
where $p^t_\tau(y,x) = \mathbb{P}(x_{t+\tau}=x | x_t=y)$ is a transition density describing the probability that a system at $y$ at time $t$ evolves to $x$ after a lag time $\tau$.  In general,  $p^t_\tau(y,x)$ depends on not only current state $y$ at time $t$, but also previous history, and is therefore time dependent.  Under the Markovian assumption, which becomes an increasingly good approximation at larger lag times $\tau$, $p^t_\tau(y,x)$ becomes a time homogeneous transition density $p_\tau(y,x)$ independent of $t$ and the transfer operator $\mathscr{T}_t$ can be written as $\mathscr{T}$, where
\begin{equation}\label{transferhomo}
	u_{t+\tau}(x)=\mathscr{T} \circ u_t(x)=\frac{1}{\pi(x)} \int dy\ p_\tau(y,x)u_t(y)\pi(y).
\end{equation}	
If the system is at equilibrium, then it additionally obeys detailed balance such that	
	\begin{equation}\label{detailed}
	\pi(x)p_\tau(x,y)=\pi(y)p_\tau(y,x).
	\end{equation}
Given any two state functions $u_1(x)$ and $u_2(x)$, we appeal to \blauw{Eq.~\ref{transferhomo} and \ref{detailed}} to write
	\begin{align*}\label{key}
&	\langle u_1(x)\vert \mathscr{T}\circ u_2(x)\rangle_\pi \\
=& \int dx\ u_1(x)\int dy\ p_\tau(y,x)u_2(y)\pi(y)\\
=& \int dx\ u_1(x)\int dy\ p_\tau(x,y)u_2(y)\pi(x)\\
=& \int dy\ u_2(y) \int dx\ p_\tau(x,y)u_1(x)\pi(x)\\
=& \int dx\ u_2(x) \int dy\ p_\tau(y,x)u_1(y)\pi(y)\\
=  &\langle \mathscr{T}\circ u_1(x)\vert  u_2(x)\rangle_\pi,
	\end{align*}
which demonstrates that $\mathscr{T}$ is self-adjoint with respect to the inner product 	
	\begin{equation}\label{key}
	\langle a\vert b\rangle_\pi=\int a(x)b(x)\pi(x)dx.
	\end{equation}
	
Let $\{\psi_i(x)\}$ be eigenfunctions of $\mathscr{T}$ corresponding to the eigenvalues $\{\lambda_i\}$ in non-ascending order 
	\begin{equation}\label{key}
	\mathscr{T} \circ \psi_i(x)=\lambda_i \psi_i(x).
	\end{equation}
	
	\noindent The self-adjoint nature of $\mathscr{T}$ implies that it possesses real eigenvalues $\{\lambda_i(x)\}$ and its eigenvectors $\{\psi_i(x)\}$ form a complete orthonormal basis~\cite{wu2017variational, nuske2014variational, noe2013variational}, with orthonormality relations
	\begin{equation}\label{ortho}
	\langle \psi_i\vert \psi_j\rangle_\pi=\delta_{ij}.
	\end{equation}
Normalization of the transition density $\int dx p_\tau(y,x) = 1$ together with the assumption of ergodicity implies that the eigenvalue spectrum is bounded from above by a unique unit eigenvalue such that $1 = \lambda_0 > \lambda_1 \geq \lambda_2 \geq \ldots$ \cite{nuske2014variational, prinz2011markov}. Any state $\chi_t(x)$ at a specific time $t$ can be written as a linear expansion in this basis of $\{\psi_i\}$
	\begin{equation}\label{key}
	\chi_t(x) = \sum_i  \langle \psi_i\vert \chi_t\rangle_\pi \psi_i(x).
	\end{equation}
The evolution of $\chi_t(x)$ to $\chi_{t+k\tau}(x)$ after a time period $k \tau$ can be written as 
	\begin{align}\label{evol}
	\chi_{t+k\tau}(x) = \mathscr{T}^k \circ \chi_t(x) &= \sum_i  \langle \psi_i\vert \chi_t\rangle_\pi \mathscr{T}^k \psi_i(x)\nonumber\\
	&= \sum_i  \langle \psi_i\vert \chi_t\rangle_\pi \lambda_i^k \psi_i(x)\nonumber\\
	&=\sum_i  \langle \psi_i\vert \chi_t\rangle_\pi \exp\p{-\frac{k\tau}{t_i}} \psi_i(x),
	\end{align}
where $t_i$ is the implied timescale corresponding to eigenfunction $\psi_i$ given by
	\begin{equation}\label{timescale}
	t_i=-\frac{\tau}{\log \lambda_i}.
	\end{equation}
	

\rood{This development makes clear that the eigenvalue associated with an eigenfunction characterizes its temporal autocorrelation. The pair $\{ \psi_0, \lambda_0 = 1\}$ corresponds to the equilibrium distribution. Smaller positive eigenvalues in the range $1 > \lambda > 0$ decay increasingly faster in time. The self-adjointness of $\mathscr{T}$ assure all eigenvalues are real but does not prohibit negative eigenvalues, which are mathematically admissible but unphysical on the grounds that they are measures of autocorrelation. Accordingly, negative eigenvalues are rarely observed for well trained models for which sufficient training data is available at sufficiently high temporal resolution, and their appearance is a numerical indication that the ``slow modes'' identified by the approach cannot be adequately resolved from the data and/or model at hand.}

\subsection{Variational approach to conformational dynamics (VAC)}

	Under the VAC, we seek an orthonormal set $\{\tilde{\psi}_i\}$ to approximate $\{\psi_i\}$ under the orthogonality conditions given by \blauw{Eq.~\ref{ortho}}. Typically we are interested not in the full $\{\psi_i\}$ but only the leading eigenfunctions corresponding to the largest eigenvalues and thus longest implied timescales. 
	
	We first observe that $\psi_0(x) = 1$ is a trivial eigenfunction of $\mathscr{T}$ with eigenvalue $\lambda_0 = 1$ corresponding to the equilibrium distribution at $t \rightarrow \infty$. This follows from \blauw{Eq.~\ref{transferhomo} and \ref{detailed}} whereby $\mathscr{T} \circ \psi_0(x) = \mathscr{T} \circ 1 = \frac{1}{\pi(x)} \int dy\ p_\tau(y,x)\pi(y) = \frac{1}{\pi(x)} \int dy\ p_\tau(x,y)\pi(x) = 1 = \psi_0(x)$.
	
	To learn $\psi_1(x)$, we note that any state function $u(x)$ which is orthogonal to $\psi_0(x)$ can be expressed as
	\begin{equation}\label{key}
u(x) = \sum_{i\ge 1}  \langle \psi_i\vert u\rangle_\pi \psi_i(x)= \sum_{i\ge 1}  c_i \psi_i(x)
	\end{equation}
where $c_i = \langle \psi_i\vert u\rangle_\pi$ are expansion coefficients, and
	\begin{equation}\label{eigenvalue_1}
	\tilde{\lambda} = \frac{\bk{u}{\mathscr{T}\circ u}_\pi}{\bk{u}{u}_\pi}=\frac{\sum_{i\ge 1} c_i^2 \lambda_i}{\sum_{i\ge 1} c_i^2}\le \frac{\sum_{i\ge 1} c_i^2 \lambda_1}{\sum_{i\ge 1} c_i^2} = \lambda_1.
	\end{equation}
Since $\tilde{\lambda}$ is bounded from above by $\lambda_1$ by the variational principle \cite{nuske2014variational, schwantes2015modeling}, we can exploit this fact to approximate the first non-trivial eigenfunction $\psi_1$ by searching for a $u$ that maximizes $\tilde{\lambda}$ subject to $\bk{u}{\psi_0}_\pi=0$.  The learned $u$ is an approximation to first non-trivial eigenfunction $\tilde{\psi}_1$. 

We can continue this procedure to approximate higher order eigenfunctions. In general we approximate $\tilde{\psi}_i(x)$ by maximizing
\begin{equation}\label{eigenvalue}
	\tilde{\lambda}_i= \frac{\bk{\tilde{\psi}_i}{\mathscr{T}\circ \tilde{\psi}_i}_\pi}{\bk{\tilde{\psi}_i}{\tilde{\psi}_i}_\pi},
	\end{equation}
under the orthogonality constraints
\begin{equation}\label{ortho_constraint}
	\bk{\tilde{\psi}_k}{\tilde{\psi}_i}_\pi=0,  \  0\le k < i.
\end{equation}

In essence, the VAC procedure combines a variational principle \cite{nuske2014variational} with a linear variational approach perhaps most familiar from quantum mechanics \cite{szabo2012modern}. Given an arbitrary input basis $\{\zeta_j(x)\}$, the eigenfunction approximations may be written as linear expansions
\begin{equation}
\tilde{\psi}_i = \sum_j s_{ij} \zeta_j.
\end{equation}
Adopting this basis, the VAC can be shown to lead to a generalized eigenvalue problem analogous to the quantum mechanical Roothaan-Hall equations \cite{nuske2014variational, szabo2012modern}
\begin{equation}\label{geneig}
C s_i = \tilde{\lambda}_i Q s_i,
\end{equation}
where
\begin{equation}\label{geneig_c}
C_{jk} = \langle \zeta_j(x)\vert \mathscr{T}\circ \zeta_k(x)\rangle_\pi,
\end{equation}
\begin{equation}\label{geneig_q}
Q_{jk} = \langle \zeta_j(x)\vert \zeta_k(x)\rangle_\pi.
\end{equation}
Here $s_i$ is the (eigen)vector of linear expansion coefficients for the approximate eigenfunction $\tilde{\psi}_i$, and $\tilde{\lambda}_i$ is the associated eigenvalue. The spectrum of solutions of \blauw{Eq.~\ref{geneig}} yield the best linear estimations of the eigenfunctions of the transfer operator $\mathscr{T}$ within the basis $\{\zeta_j(x)\}$. The generalized eigenvalue problem can be solved by standard techniques \cite{watkins2007matrix}.

\blauw{Eq.~\ref{eigenvalue} and \ref{ortho_constraint}} serve as the central equations for TICA, kTICA and SRVs. We first show how these equations can be estimated from simulated data, and then how these three methods emerge as specializations of VAC under particular choices for the input basis.

\subsection{Estimation of VAC equations from trajectory data}

Here we show how \blauw{Eq.~\ref{eigenvalue} and \ref{ortho_constraint}} can be estimated from empirical trajectory data \cite{noe2013variational, nuske2014variational, wu2017variational}. The numerator of \blauw{Eq.~\ref{eigenvalue}} becomes 
\begin{align}\label{key}
&	\bk{\tilde{\psi}_i}{\mathscr{T}\circ \tilde{\psi}_i}_\pi \nonumber \\
=& \int dx\ \pi(x)\tilde{\psi}_i(x)\frac{1}{\pi(x)}\int dy\ p_\tau(y,x)\tilde{\psi}_i(y)\pi(y) \nonumber \\
=&\int dx\ dy\ \tilde{\psi}_i(x) p_\tau(y,x) \tilde{\psi}_i(y) \pi(y) \nonumber \\
=&\int dx\ dy\ \tilde{\psi}_i(x) \mathbb{P}(x_{t+\tau}=x | x_t=y) \tilde{\psi}_i(y) \mathbb{P}(x_t=y) \nonumber \\
\approx & \, \Expect{\tilde{\psi}_i(x_t)\tilde{\psi}_i(x_{t+\tau})},
\end{align}
where $\Expect{\tilde{\psi}_i(x_t)\tilde{\psi}_i(x_{t+\tau})}$ can be estimated from a trajectory $\{x_t\}$. The denominator follows similarly as 
\begin{align}
	&	\bk{\tilde{\psi}_i}{ \tilde{\psi}_i}_\pi \nonumber \\
	=&\int dx\ \pi(x)\tilde{\psi}_i(x)\tilde{\psi}_i(x) \nonumber \\
	=&\int dx\ \tilde{\psi}_i(x)\tilde{\psi}_i(x) \mathbb{P}(x_t=x) \nonumber \\
	\approx & \, \Expect{\tilde{\psi}_i(x_t)\tilde{\psi}_i(x_{t})}.
\end{align}
The full expression for \blauw{Eq.~\ref{eigenvalue}} becomes
\begin{equation}\label{eigenvalue_estimation}
	\tilde{\lambda}_i= \frac{\bk{\tilde{\psi}_i}{\mathscr{T}\circ \tilde{\psi}_i}_\pi}{\bk{\tilde{\psi}_i}{\tilde{\psi}_i}_\pi}\approx\frac{\Expect{\tilde{\psi}_i(x_t)\tilde{\psi}_i(x_{t+\tau})}}{\Expect{\tilde{\psi}_i(x_t)\tilde{\psi}_i(x_{t})}}.
\end{equation}
Similarly, \blauw{Eq.~\ref{ortho_constraint}} becomes
\begin{equation}\label{key}
	\bk{\tilde{\psi}_k}{\tilde{\psi}_i}_\pi\approx\Expect{\tilde{\psi}_k(x_t)\tilde{\psi}_i(x_{t})}=0,  \  0\le k < i.
\end{equation}
Using the same reasoning, the components (\blauw{Eq.~\ref{geneig_c} and \ref{geneig_q}}) of the generalized eigenvalue problem (\blauw{Eq.~\ref{geneig}}) are estimated as

	\begin{equation}\label{geneig_c_e}
	C_{jk} = \langle \zeta_j(x)\vert \mathscr{T}\circ \zeta_k(x)\rangle_\pi \approx \Expect{\zeta_j(x_t)\zeta_k(x_{t+\tau})},
	\end{equation}
	\begin{equation}\label{geneig_q_e}
	Q_{jk} = \langle \zeta_j(x)\vert \zeta_k(x)\rangle_\pi\approx \Expect{\zeta_j(x_t)\zeta_k(x_{t})}.
	\end{equation}

	\subsection{Time-lagged independent component analysis (TICA)}
	
	In TICA, we represent $\tilde{\psi}_i(x)$ as a linear combination of molecular coordinates 
	$x$, where $a_i$ is a vector of linear expansion coefficients and $C$ is an additive constant 
	
	\begin{equation}\label{eq_u}
	\tilde{\psi}_i(x) = a_i\cdot  x + C.
	\end{equation}
	
	\noindent The orthogonality condition \blauw{Eq.~\ref{ortho_constraint}} of $\tilde{\psi}_i$ relative to $\tilde{\psi}_0(x) = 1$ becomes
	
\begin{equation}\label{key}
	0=\int dx\ \pi(x)\tilde{\psi}_0(x)\tilde{\psi}_i(x)=\int dx\ \pi(x)\tilde{\psi}_i(x)=\Expect{\tilde{\psi}_i(x)}.
\end{equation}
It follows that
\begin{align}\label{key}
	0&=\Expect{\tilde{\psi}_i(x)}=\Expect{a_i\cdot  x + C} = a_i \cdot \Expect{x} + C \\
	\Rightarrow C &= -a_i \cdot \Expect{x},
\end{align}
and therefore \blauw{Eq.~\ref{eq_u}} can be written as
\begin{equation}\label{key}
	\tilde{\psi}_i(x) = a_i\cdot x - a_i  \cdot \Expect{ x} = a_i \cdot  \delta x,
\end{equation}
where $\delta x = x-\Expect{x}$ is a mean-free coordinate. Under this specification for $\tilde{\psi}_i(x)$,  \blauw{Eq.~\ref{eigenvalue}} and \blauw{Eq.~\ref{ortho_constraint}} become
	
	\begin{align}\label{eq11}
	\tilde{\lambda}_i &= \frac{\bk{\tilde{\psi}_i}{\mathscr{T}\circ \tilde{\psi}_i}_\pi}{\bk{\tilde{\psi}_i}{\tilde{\psi}_i}_\pi}\nonumber \\
	&= \frac{\Expect{(a_i\cdot \delta x_t)(a_i\cdot \delta x_{t+\tau})}}{\Expect{(a_i\cdot \delta x_t)^2}},
	\end{align}
	
	\begin{equation}\label{eq12}
	0 = \rood{\bk{\tilde{\psi}_k}{\tilde{\psi}_i}_\pi} = \Expect{(a_i\cdot \delta x_t)(a_k\cdot \delta x_{t})}.
	\end{equation}
	
\noindent which are exactly the objective function and orthogonality constraints of TICA \cite{schwantes2013improvements, perez2013identification}.

\subsection{Kernel TICA (kTICA)}
	
	One way of generalizing TICA to learn nonlinear features is though feature engineering. Specifically, if we can find a 
	nonlinear mapping $\phi$ that maps configurations $x$ to appropriate features $\phi(x)$, we can apply TICA on these
	features. However, designing good nonlinear features typically requires expert knowledge or expensive data preprocessing
	techniques. Therefore, instead of finding an explicit mapping $\phi$, an  alternative approach is to apply the kernel trick using a
	kernel function $K(x,y) = \phi(x) \cdot \phi(y)$ that defines an inner product between $\phi(x)$ and $\phi(y)$ as a similarity measure in the feature space that does not require explicit definition of $\phi$. 
	
	
	To apply the kernel trick to TICA, we need to reformulate TICA in terms of this kernel function. It can be shown that in \blauw{Eq.~\ref{eq11}}, the coefficient $a_i$ is linear combination of $\{\delta x_t\} \cup \{\delta x_{t+\tau}\}$ (see Supplementary Information of Ref.~\cite{schwantes2015modeling}) and may therefore be written as
	\begin{equation}\label{ai}
	a_i=\sum_t\p{\beta_{it} \delta x_t + \gamma_{it} \delta x_{t+\tau}}.
	\end{equation}	
Under this definition for $a_i$ \blauw{Eq.~\ref{eq11} and \ref{eq12}} become,
\begin{widetext}
	
	\begin{align}\label{eq11.1}
	\tilde{\lambda}_i &= \frac{\bk{\tilde{\psi}_i}{\mathscr{T}\circ \tilde{\psi}_i}_\pi}{\bk{\tilde{\psi}_i}{\tilde{\psi}_i}_\pi}\nonumber \\ 
	&= \dfrac{
		\Expect{
			(\sum_t\p{\beta_{it} \delta x_t + \gamma_{it} \delta x_{t+\tau}}\cdot \delta x_t)
			(\sum_t\p{\beta_{it} \delta x_t + \gamma_{it} \delta x_{t+\tau}}\cdot \delta x_{t+\tau})   
	}}{\Expect{(\sum_t\p{\beta_{it} \delta x_t + \gamma_{it} \delta x_{t+\tau}}\cdot \delta x_t)^2}},
	\end{align}

	\begin{align}\label{eq12.1}
	0 &= \bk{\tilde{\psi}_k}{\tilde{\psi}_i}_\pi\nonumber \\
	&= \Expect{
		{(\sum_t\p{\beta_{it} \delta x_t + \gamma_{it} \delta x_{t+\tau}}\cdot \delta x_t)}
		{(\sum_t\p{\beta_{kt} \delta x_t + \gamma_{kt} \delta x_{t+\tau}}\cdot \delta x_{t})}
	}.
	\end{align}
	
\end{widetext}
Now the objective function \blauw{Eq.~\ref{eq11.1}} and constraints \blauw{Eq.~\ref{eq12.1}} only depend on 
	inner products between any pair of elements in $\{\delta x_t\} \cup \{\delta x_{t+\tau}\}$.
		
To obtain a nonlinear transformation, we replace the linear similarity measure, which is the inner product 
	$\delta x \cdot \delta y$ of two vectors, with a symmetric nonlinear kernel function $K(x,y)$.  
	This transforms \blauw{Eq.~\ref{eq11.1} and \ref{eq12.1}} to
\begin{widetext}
	\begin{align}\label{eq11.2}
	\tilde{\lambda}_i &= \frac{\bk{\tilde{\psi}_i}{\mathscr{T}\circ \tilde{\psi}_i}_\pi}{\bk{\tilde{\psi}_i}{\tilde{\psi}_i}_\pi}\nonumber \\
	&= \frac{\Expect{
			{(\sum_t\p{\beta_{it} K(x_t,x_t) + \gamma_{it} K(x_t,x_{t+\tau})})}
			{(\sum_t\p{\beta_{it} K(x_t, x_{x+\tau}) + \gamma_{it} K(x_{t+\tau}, x_{t+\tau})})}
	}}{\Expect{(\sum_t\p{\beta_{it} K(x_t,x_t) + \gamma_{it} K(x_t,x_{t+\tau})})^2}},
	\end{align}
	\begin{align}\label{eq12.2}
	0 &= \bk{\tilde{\psi}_k}{\tilde{\psi}_i}_\pi\nonumber \\ 
	&= \Expect{ 
		{(\sum_t\p{\beta_{it} K(x_t,x_t) + \gamma_{it} K(x_t,x_{t+\tau})})}
		{(\sum_t\p{\beta_{kt} K(x_t,x_t) + \gamma_{kt} K(x_t,x_{t+\tau})})}
	},
	\end{align}
\end{widetext}
	\noindent which define the objective function and orthogonality constraints of kTICA. As detailed in Ref.~\cite{schwantes2015modeling}, \blauw{Eq.~\ref{eq11.2} and \ref{eq12.2}} can be simplified to a generalized eigenvalue problem that admits efficient solution by standard techniques \cite{watkins2007matrix}.
	
	Although kTICA enables recovery of nonlinear eigenfunctions, it does have some
	significant drawbacks.  First, it has high time and space complexity. The Gram matrix $K=[K(x, y)]_{N\times N}$ takes 
	$O(N^2)$ time to compute and requires $O(N^2)$ memory, and the generalized eigenvalue problem takes $O(N^3)$ time to solve, which
	severely limits its application to large datasets. Second, results can be sensitive to the choice of kernel and there exist no rigorous guidelines for the choice of an appropriate kernel for a particular application. This limits the generalizability of the approach. Third, the kernels are typically highly sensitive to the choice of hyperparameters. For example, the Gaussian (radial basis function) kernel
	is sensitive to noise for small kernel widths $\sigma$, which leads to overfitting and overestimation the implied timescales.
	A large $\sigma$ on the other hand typically approaches linear TICA results which undermines its capacity to learn nonlinear
	transformations \cite{harrigan2017landmark, sultan2018transferable}. This hyperparameter sensitivity typically requires signifiant expenditure of computational effort to tune these values in order to obtain satisfactory results \cite{schwantes2015modeling, harrigan2017landmark}. Fourth, kTICA does not furnish an explicit expression for the mapping $\phi: x \rightarrow \phi(x)$ projecting configurations into the nonlinear feature space \cite{hernandez2018variational}. Accordingly, it is not straightforward to apply kernel TICA within enhanced sampling protocols that require the learned latent variables to be explicit and differentiable functions of the input coordinates.
	
	One way to ameliorate the first deficiency by reducing memory usage and computation time is to employ a variant of kTICA known as landmark kTICA. This approach selects $m<<N$ landmarks from the dataset, computes an 
	$m$-by-$m$ Gram matrix, and then uses the Nystr\"om approximation to estimate the original $N$-by-$N$ Gram 
	matrix~\cite{harrigan2017landmark}. 
	
	\subsection{\rood{State-free reversible VAMPnets (SRVs)}}
	
	In general, \rood{the eigenfunction approximations} need not be a linear function within either the input coordinate space or the feature space. The most general form of \blauw{Eq.~\ref{eigenvalue} and \ref{ortho_constraint}} is
	
	\begin{align}\label{eq13}
	\tilde{\lambda}_i &= \frac{\bk{\tilde{\psi}_i}{\mathscr{T}\circ \tilde{\psi}_i}_\pi}{\bk{\tilde{\psi}_i}{\tilde{\psi}_i}_\pi}\nonumber \\  
	&= \frac{\Expect{\tilde{\psi}_i(x_t)\tilde{\psi}_i(x_{t+\tau})}}{\Expect{\tilde{\psi}_i^2(x_t)}},
	\end{align}
	
	\begin{equation}\label{eq14}
	0 = \bk{\tilde{\psi}_k}{\tilde{\psi}_i}_\pi = \Expect{\tilde{\psi}_i(x_t)\tilde{\psi}_k(x_t)}.
	\end{equation}
	
	We now introduce the SRV approach that employs a neural network $f$ to learn nonlinear approximations to $\{\tilde{\psi}_i\}$ directly without requiring the kernel trick. The neural network $f$ maps configuration $x$ to a $n$-dimensional output $f_i(x)\ (i=1,...,n)$, where $n$ is the
	number of slow modes we want to learn. Then a linear variational method is applied to obtain the corresponding $\tilde{\psi}_i(x)$ such that 
	$\{\tilde{\psi}_i(x)\}$ form an orthonormal basis set that minimizes the network loss function. 
	
	The method proceeds as follows. Given a neural network $f$ with $n$-dimensional outputs, we feed a training set $X=\{x\}$ and
	train the network with loss function $L$
	\begin{equation}\label{loss_f}
	L = \sum_i g(\tilde{\lambda}_i)=\sum_i g\p{\frac{\Expect{\tilde{\psi}_i(x_t)\tilde{\psi}_i(x_{t+\tau})}}{\Expect{\tilde{\psi}_i^2(x_t)}}}.
	\end{equation}
	where $g$ is a monotonically decreasing function. Minimizing $L$ is equivalent to maximizing the sum over $ \tilde{\lambda}_i$ and therefore maximizing the sum over $t_i$ (\blauw{Eq.~\ref{timescale}}). The $\{\tilde{\psi}_i\}$ correspond to linear combinations of the neural network outputs $\{f_i(x)\}$ 
	\begin{equation}\label{ui}
	\tilde{\psi}_i(x) = \sum_j s_{ij} f_j(x),
	\end{equation}
computed by applying the linear VAC to the neural network outputs~\cite{noe2013variational, nuske2014variational}.  The linear VAC is equivalent to the generalized eigenvalue problem in \blauw{Eq.~\ref{geneig}} where $\zeta_j(x)=f_j(x)$. 

Minimization of the loss function by gradient descent requires the derivative of $L$ with respect to neural network parameters $\theta$
\begin{equation}\label{key}
\pd{L}{\theta}=\sum_i \pd{L}{g}\pd{g}{\tilde{\lambda}_i}\pd{\tilde{\lambda}_i}{\theta}.
\end{equation}
The first two partial derivatives are straightforwardly calculated by automatic differentiation once a choice for $g$ has been made. The third partial derivative requires a little more careful consideration. We first expand this final derivative using \blauw{Eq.~\ref{geneig}} to make explicit the dependence of $\tilde{\lambda}_i$ on the matrices $C$ and $Q$ 
\begin{equation}\label{difficult_diff}
\pd{\tilde{\lambda}_i}{\theta}=\pd{\tilde{\lambda}_i}{C}\pd{C}{\theta}+\pd{\tilde{\lambda}_i}{Q}\pd{Q}{\theta}.
\end{equation}
To our best knowledge, no existing computational graph frameworks provide gradients for generalized eigenvalue problems. Accordingly, we rewrite \blauw{Eq.~\ref{geneig}} as follows. We first apply a Cholesky decomposition to $Q$, such that
\begin{equation}\label{key}
C s_i = \tilde{\lambda}_i LL^T s_i,
\end{equation}
where $L$ is a lower triangular matrix. We then left multiply both sides by $L^{-1}$ to obtain
\begin{equation}\label{key}
\p{L^{-1}C (L^{T})^{-1}} \p{L^T s_i}  = \tilde{\lambda}_i \p{L^T s_i}.
\end{equation}
Defining $\tilde{C} = L^{-1}C (L^{T})^{-1}$ and $\tilde{s}_i=L^T s_i$ we convert the generalized eigenvalue problem into a standard eigenvalue with a symmetric matrix $\tilde{C}$
\begin{equation}\label{key}
\tilde{C} \tilde{s}_i  = \tilde{\lambda}_i \tilde{s}_i ,
\end{equation}
where the Cholesky decomposition assures numerical stability. Now, \blauw{Eq.~\ref{difficult_diff}} becomes
\begin{equation}\label{key}
\pd{\tilde{\lambda}_i}{\theta}=\pd{\tilde{\lambda}_i}{\tilde{C}} \p{\pd{\tilde{C}}{C}\pd{C}{\theta}+\pd{\tilde{C}}{Q}\pd{Q}{\theta}}
\end{equation}
where all terms are computable using automatic differentiation: $\pd{\tilde{\lambda}_i}{\tilde{C}}$ from routines for a symmetric matrix eigenvalue problem, $\pd{\tilde{C}}{C}$ and $\pd{\tilde{C}}{Q}$ from those for Cholesky decomposition, matrix inversion, and matrix multiplication, and $\pd{C}{\theta}$ and $\pd{Q}{\theta}$ by applying the chain rule to \blauw{Eq.~\ref{geneig_c} and \ref{geneig_q}} with $\zeta_j(x)=f_j(x)$ and computing the derivatives $\pd{f}{\theta}$ through the neural network.

\begin{figure}[ht!]
		\begin{center}
			\includegraphics[width=\figurewidth]{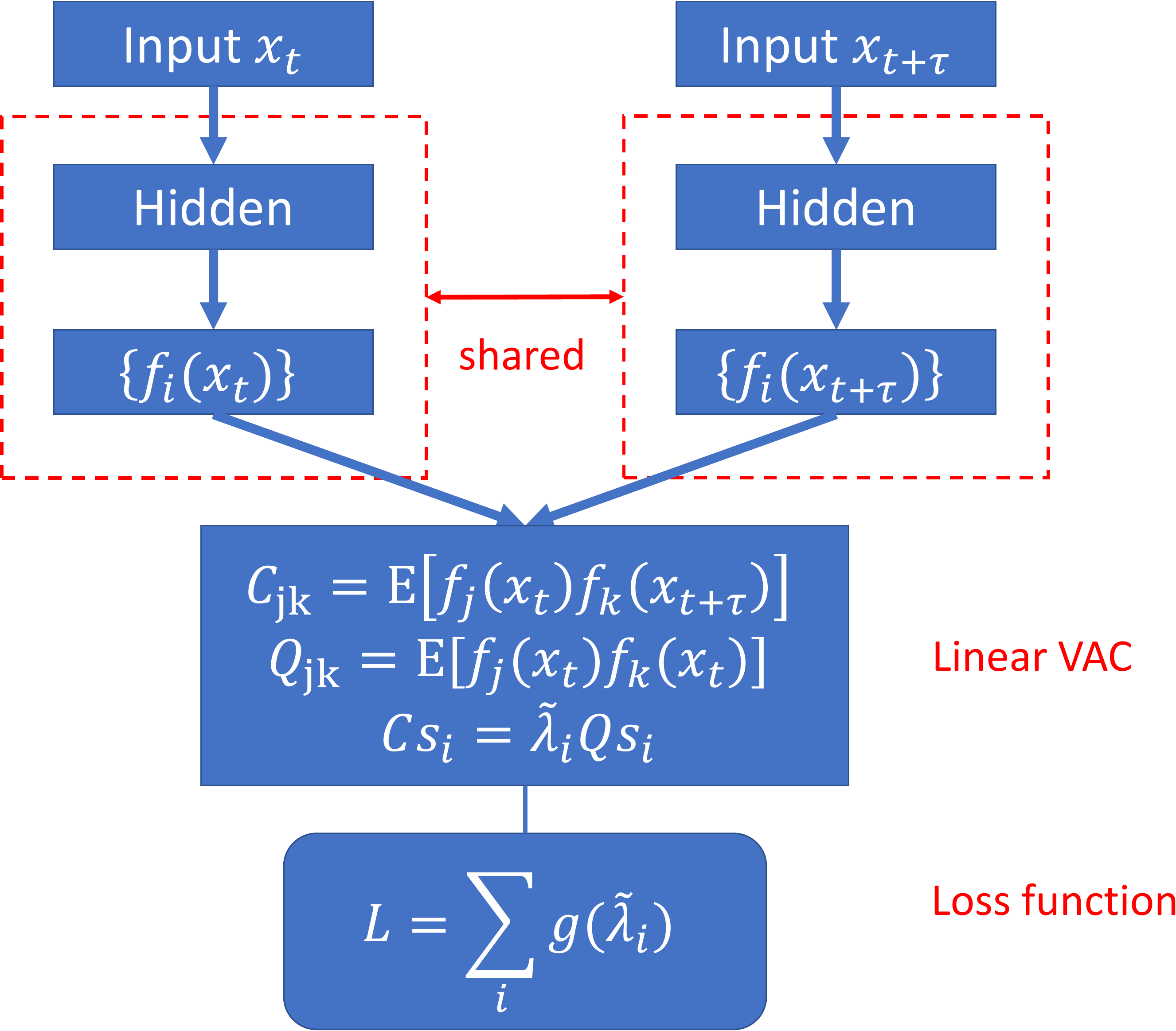}
			\caption{
			Schematic diagram of \rood{state-free reversible VAMPnets (SRVs)}. A pair of configurations $(x_t,x_{t+\tau})$ are fed into a Siamese neural network in which each subnet possesses the same architecture and weights. The Siamese net generates network outputs $\{f_i(x_t)\}$ and $\{f_i(x_{t+\tau})\}$ . The mappings $f_i: x \rightarrow f_i(x)$ evolve as network training proceeds and the network weights are updated.  Optimal linear combinations of $\{f_i\}$ produce estimates of the transfer operator eigenfunctions $\{\tilde{\psi}_i\}$ by applying linear VAC, which can be formulated as a generalized eigenvalue problem. Network training proceeds by backpropagation and is terminated when the loss function $L=\sum_ig(\tilde{\lambda}_i)$ is minimized. }
			\label{SRV_diagram}
		\end{center}
	\end{figure}

Training is performed by passing $\{x_t,x_{t+\tau}\}$ pairs to the SRV and updating the network parameters using mini-batch gradient descent using Adam~\cite{kingma2014adam} and employing the automatic differentiation expression for $\partial L / \partial \theta$ to minimize the loss function. To prevent overfitting, we shuffle all $\{x_t,x_{t+\tau}\}$ pairs and reserve a small portion as validation set with which to implement early stopping.  Training is terminated when validation loss no longer decreases for a pre-specified number of epochs. A schematic diagram of the SRV is shown in \blauw{Fig.~\ref{SRV_diagram}}.

	We note that we do not learn $\{\tilde{\psi}_i\}$ directly in our neural network, but obtain it as a weighted linear combination of $\{f_i\}$.  
	Specifically, during training we learn not only the weights of neural network, but also the linear expansion coefficients $\{s_{ij}\}$ 
	that yield $\{\tilde{\psi}_i\}$ from $\{f_i\}$. Conceptually, one can consider the neural network as learning an optimal nonlinear featurization of the input basis to pass through the VAC framework. After training is complete, a new out-of-sample configuration $x$ can be passed through the neural network to produce $f$, which is then transformed via a linear operator $\{s_{ij}\}$ to get the 
	eigenfunctions estimates $\{\tilde{\psi}_i\}$. Since the neural network is fully differentiable and the final transformation is linear the SRV mapping $x \rightarrow \{\tilde{\psi}_i\}$ is explicit and differentiable, making it well suited to applications in enhanced sampling.

	An important choice in our network design is the function $g(\tilde{\lambda})$ within our loss function. In theory it can be any monotonically decreasing function,
	but motivated by Refs.~\cite{mardt2018vampnets} and \cite{wu2017variational}, we find that choosing $g(\tilde{\lambda})=-\tilde{\lambda}^2$ such that the loss function corresponds to the VAMP-2 score yields good performance and possesses strong theoretical grounding. Specifically, the VAMP-2 score may be interpreted as the cumulative kinetic variance \cite{noe2015kinetic,husic2018markov} analogous to the cumulative explained variance in principal component analysis \cite{pearson1901liii}, but where the VAMP-2 score measures kinetic rather than conformational information. The VAMP-2 score can also be considered to measure the closeness of the approximate eigenfunctions to the true eigenfunctions of the transfer operator, with this score achieving its maximum value when the approximations become exact \cite{mardt2018vampnets}.	This choice may also be generalized to the VAMP-r score $g(\tilde{\lambda})=-\tilde{\lambda}^r$ \cite{wu2017variational}. We have also observed good performance using $g(\tilde{\lambda})=1 / \log (\tilde{\lambda}$), which corresponds to maximizing the sum of 
	implied timescales $\sum_i t_i$ (\blauw{Eq.~\ref{timescale}}).

	\rood{We also note that it is possible that the transfer operator may possess degenerate eigenvalues corresponding to associated eigenfunctions with identical implied timescales. True degeneracy is expected to be rare within the leading eigenfunctions but may arise from symmetries in the system dynamics; approximate degeneracy may be encountered wherein timescales become indistinguishable within the training data available to the model. In principle, either of these situations could lead to numerical difficulties wherein degenerate eigenfunctions cannot be stably resolved due to the existence of infinitely many equivalent linear combinations. In such instances, all eigenfunctions associated with repeated eigenvalues should be retained and mutually aligned within successive rounds of training by determining the optimal rotation matrix within their linear subspace. In practice, we suspect that minor differences in the implied timescales induced by the finite nature of the training data will break the degeneracy of the eigenvalues and allow for stable numerical recovery. A \textit{post hoc} analysis of any groups of close eigenvalues can then be performed to attempt to resolve the root of any suspected underlying symmetry.}
		
	The SRV learning protocol is quite simple and efficient. It only requires $O(N)$ memory and $O(N)$ computation time, which 
	makes it ideal for large datasets. It also does not require selection of a kernel function to achieve appropriate nonlinear embeddings \cite{sultan2018transferable, hernandez2018variational} Instead we appeal to the universal approximation theorem \cite{hassoun1995fundamentals,chen1995universal}, which loosely states that a neural network with more than one hidden layer and enough number of hidden nodes can approximate any nonlinear continuous function without need to impose a kernel. Lifting the requirement for kernel selection and its attendant hyperparameter tuning is a significant advantage of the present approach over kernel-based techniques. Training such a simple neural network is possible using standard techniques such as 
	stochastic gradient descent and is largely insensitive to our choice of network architecture, hyperparameters, and activation functions. Using a default learning rate, large batch size, 
	and sufficiently large network gives excellent results as we demonstrate below. Furthermore, we use the same number of hidden layers, number of nodes in each hidden layer, and hidden layer activation functions for all four applications in this work, demonstrating the simplicity, robustness, and generalizability of the SRV approach.

	\section{Results}
	
	\subsection{1D 4-well potential}
	
\begin{figure}[ht!]
		\begin{center}
			\includegraphics[width=\figurewidth]{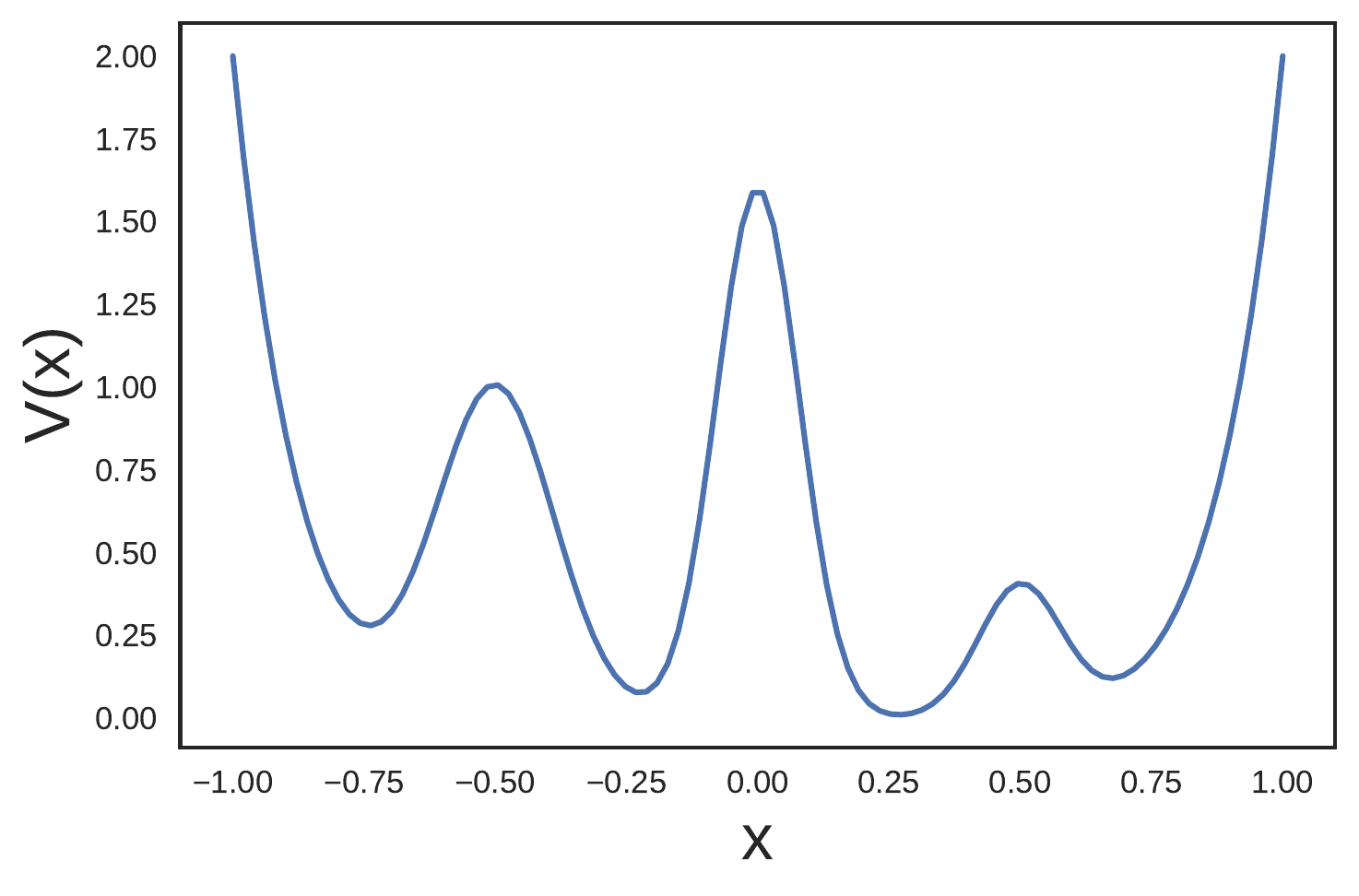}
			\caption{Model 1D 4-well potential landscape. Potential energy barriers of various heights to introduce metastability and a separation of time-scales. The potential is given by 
				$V(x) = 2\left(x^8 + 0.8e^{-80 x^2} + 0.2e^{-80(x-0.5)^2} + 0.5e^{-40(x + 0.5)^2} \right)$.} 
			\label{potential_1_fig}
		\end{center}
	\end{figure}

	In our first example, we consider a simple 1D 4-well potential defined in Ref.~\cite{schwantes2015modeling} and illustrated 
	in \blauw{Fig.~\ref{potential_1_fig}}. The eigenfunctions of the transfer operator for this system are exactly calculable, and this provides a simple initial demonstration of SRVs. We construct a transition matrix as a discrete approximation for the transfer operator by dividing the interval $[-1, 1]$ into 100 evenly-spaced bins and computing the transition probability $p_{ij}$ of moving from  bin $i$ to bin $j$ as, 

	\begin{equation}\label{transition_prop}
	p_{ij} = \begin{cases}
	C_i e^{-(V_j-V_i)}, \ \text{if } \abs{i -j} \leq 1\\
	0, \ \text{otherwise}
	\end{cases}
	\end{equation}
	
	\noindent where $V_j$ and $V_i$ are the potential energies at the centers of bins $j$ and $i$ and $C_i$ is the 
	normalization factor for bin $i$ such that the total transition probability associated with bin $i$ sums to unity. We then define a unit-time transition matrix $P(1)=[p_{ij}]_{100\times 100}$, 
	and a transition matrix of lag time $\tau$ as $P(\tau)=P(1)^\tau$. In this model, we use 
	a lag time $\tau=100$ for both theoretical analysis and model learning.
	
\begin{figure}[ht!]
		\begin{center}
			\includegraphics[width=\figurewidth]{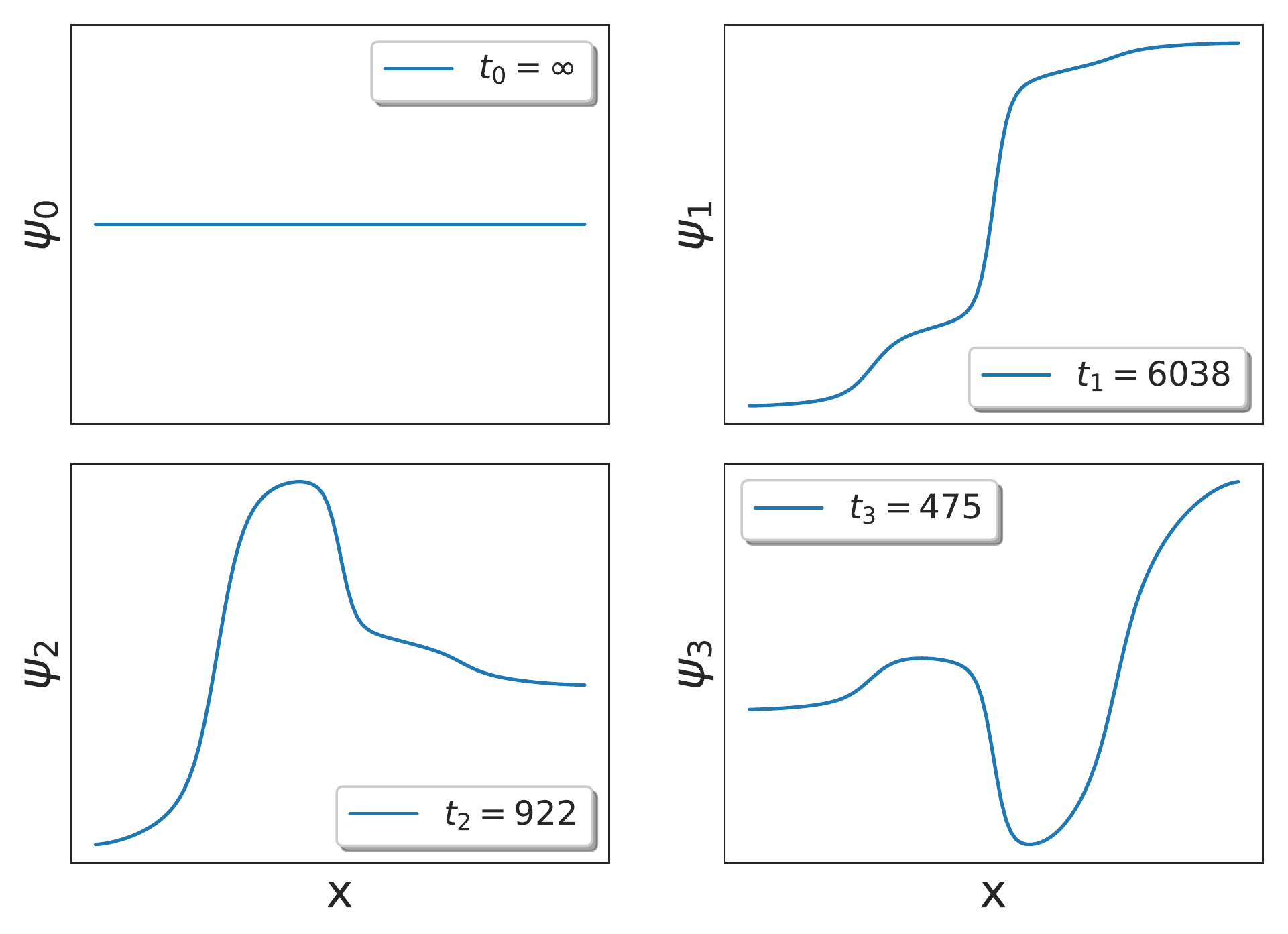}
			\caption{Theoretical eigenfunctions of the transfer operator for the model 1D 4-well potential illustrated 
				in Figure~\ref{potential_1_fig}. Note that the first eigenfunction corresponds to the stationary distribution 
				and is equal to unity for the transfer operator. The remaining eigenfunctions represent transitions between the 
				various energy wells whose barrier heights determine the associated eigenvalues $\lambda_i$ and implied timescales $t_i=-\frac{\tau}{\log \lambda_i}$. } \label{eigenstates_1}
		\end{center}
	\end{figure}

	By computing the eigenvectors $\{v_i\}$ of the transition matrix $P(\tau = 100)$, we recover the equilibrium distribution
	$\pi=v_0$. The corresponding eigenfunctions of transfer operator $\mathscr{T}(\tau=100)$ are given by $\psi_i = v_i / \pi$. 
	The top four eigenfunctions are shown in \blauw{Fig.~\ref{eigenstates_1}}, where first eigenfunction $\psi_0$ corresponds to 
	trivial stationary transition, and next three to transitions over the three energy barriers.
	
Using the calculated transition matrix $P(1)$, we generate a 5,000,000-step trajectory over the 1D landscape by initializing the system within a randomly selected bin and then propagating its dynamics forward through time under the action of $P(1)$~\cite{schwantes2015modeling}. The state of the system at any time $t$ is represented by the 1D $x$-coordinate of the particle $x(t) \in \mathbb{R}^1$. We then adopt a fixed lag time of $\tau = 100$ and learn the top three 
	non-trivial eigenfunctions using both kTICA and SRV. We note that for kTICA, we cannot compute the full 
	5,000,000-by-5,000,000 Gram matrix, so instead we select 200 landmarks by K-means clustering and use the landmark
	approximation. We select a Gaussian kernel with $\sigma=0.05$. In contrast, the SRV has no difficulty processing all data points. We use two hidden layers with 100 neurons each, giving a final architecture of [1, 100, 100, 3]. The activation function for all layers are selected to be $\tanh(x)$, and we employ a VAMP-2 loss function. The SRV network is constructed and trained within Keras~\cite{chollet2015keras}.
	
\begin{figure}[ht!]
		\begin{center}
			\includegraphics[width=\figurewidth]{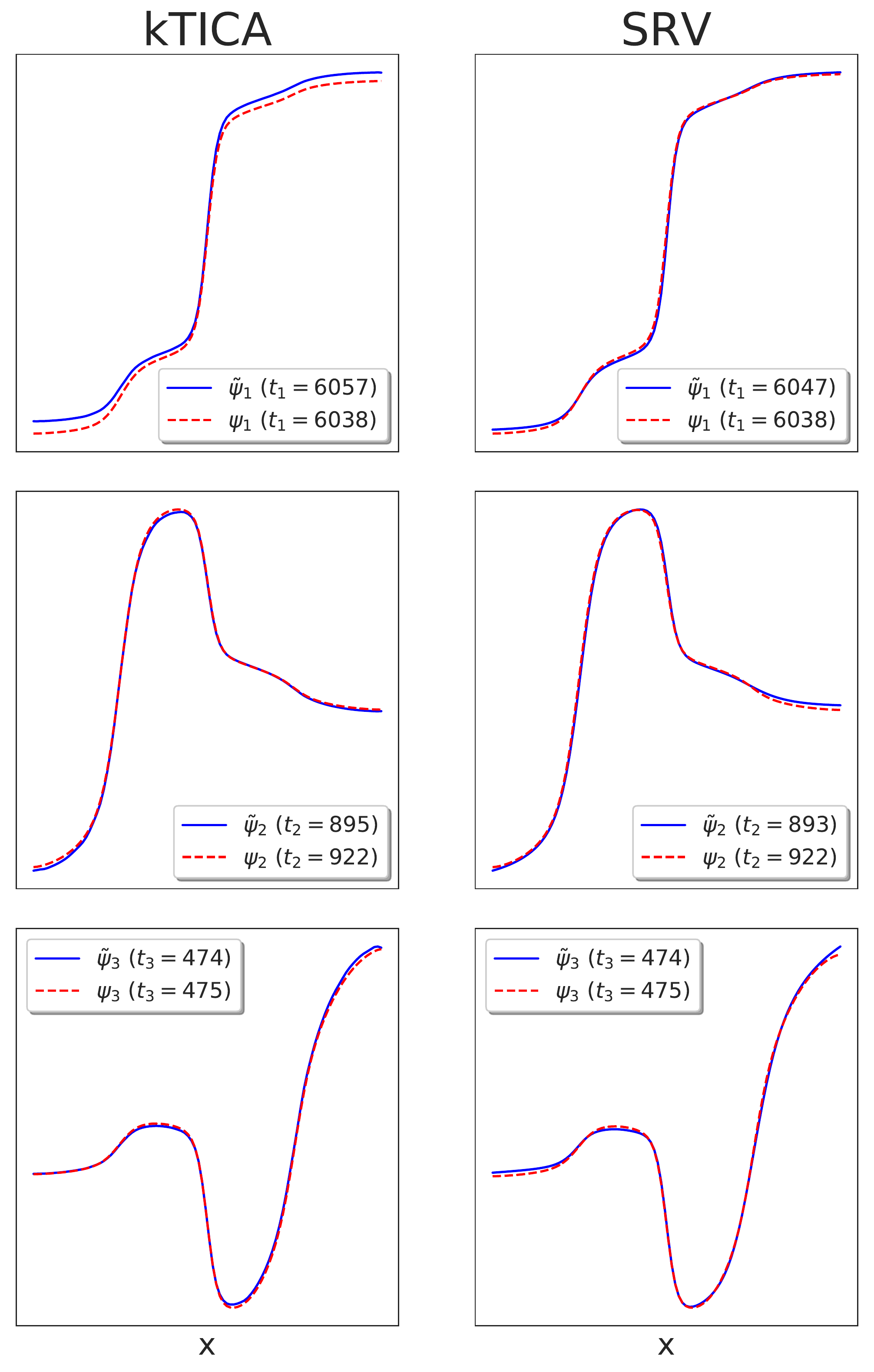}
			\caption{Top three eigenfunctions learned by kernel TICA (column 1) and SRV (column 2) for the model 1D 4-well potential. Each row represents the theoretical eigenfunction (red dashed line) and corresponding learned eigenfunction (blue solid line).  Timescales for theoretical eigenfunctions and learned eigenfunctions are reported in the legend. } \label{1D_4well_states}
		\end{center}
	\end{figure}

	The results for kTICA and SRV are shown in \blauw{Fig.~\ref{1D_4well_states}}. 
	We find that both methods are in excellent agreement with the theoretical eigenfunctions.
	The small deviation between the estimated timescales for both methods and the theoretical timescales is a 
	result of noise in the simulated data. This result demonstrates the capacity of SRVs to recover the eigenfunctions of a simple 1D system in quantitative agreement with the theoretical results and excellent correspondence with kTICA.
	
	\subsection{Ring potential}
	
\begin{figure}[ht!]
		\begin{center}
			\includegraphics[width=\figurewidth]{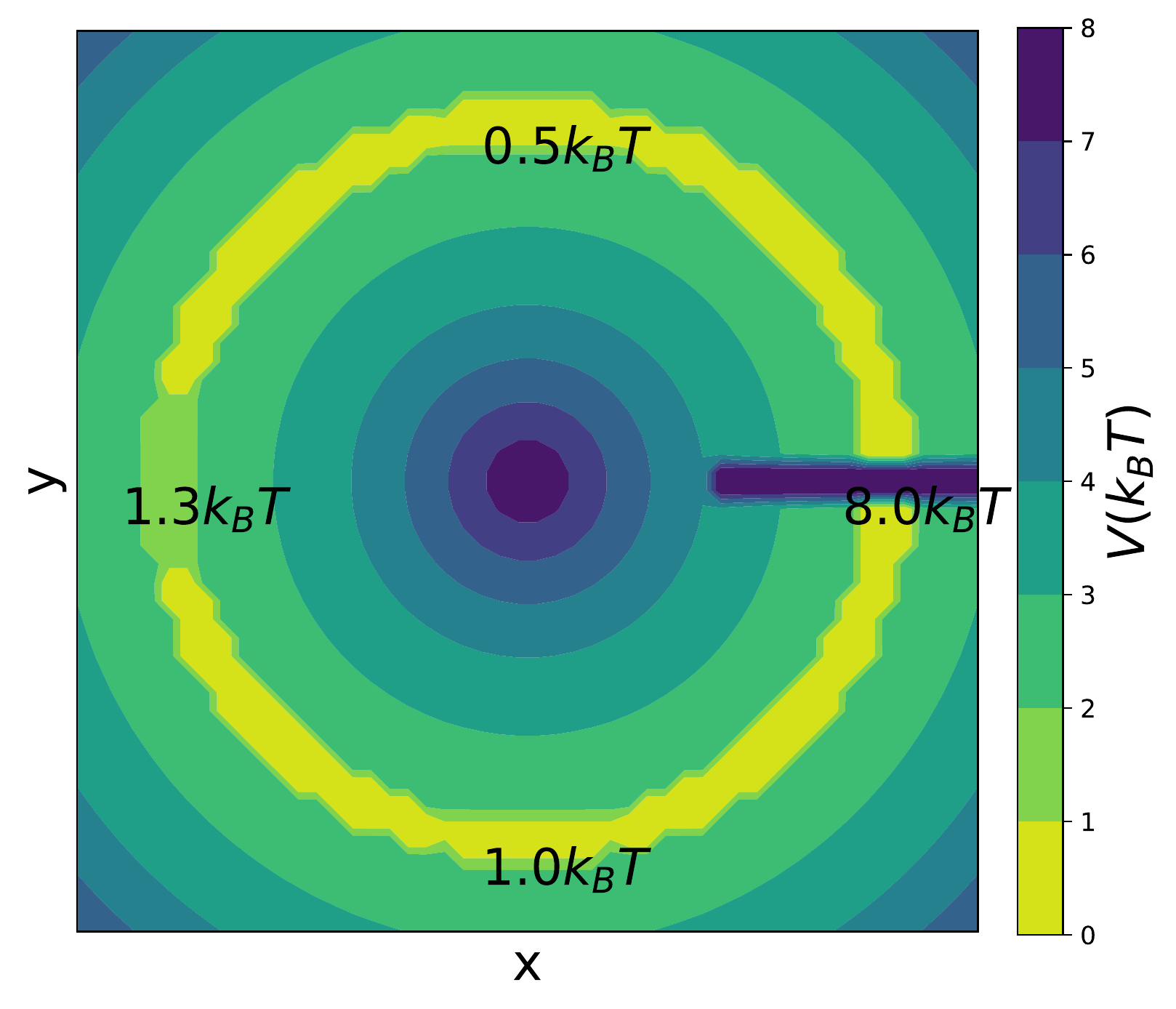}
			\caption{Contour plot of 2D ring potential, which consists a ring-shape potential valley, with four potential barriers of heights 1.0 $k_BT$, 1.3 $k_BT$, 0.5 $k_BT$, and 8.0 $k_BT$. } \label{ring_potential}
		\end{center}
	\end{figure}

	We now consider the more complicated example of a 2D modified ring potential $V(r, \theta)$.  This potential contains a narrow ring potential valley of 0 $k_BT$ and four barriers of heights 1.0 $k_BT$, 1.3 $k_BT$, 0.5 $k_BT$, and 8.0 $k_BT$. The expression of the potential is given by \blauw{Eq.~\ref{potential_ring}}, and it is plotted in  \blauw{Fig.~\ref{ring_potential}}
	
	\begin{equation}\label{potential_ring}
	\frac{V(r, \theta)}{k_BT} = \begin{cases}
	2.5 + 9(r-0.8)^2\text{,  if } \abs{r-0.8} > 0.05\\
	0.5\text{,  if } \abs{r-0.8} < 0.05 \text{ and} \abs{\theta-\frac{\pi}{2}} < 0.25\\
	1.3\text{,  if } \abs{r-0.8} < 0.05 \text{ and} \abs{\theta-{\pi}} < 0.25\\
	1.0\text{,  if } \abs{r-0.8} < 0.05 \text{ and} \abs{\theta-\frac{3\pi}{2}} < 0.25\\
	8.0\text{,  if } \abs{r} > 0.4 \text{ and} \abs{\theta} < 0.05\\
	0\text{, otherwise}
	\end{cases}
	\end{equation}

	We use the same procedure outlined in the previous example to generate the theoretical eigenfunctions of transfer operator and
	simulate a trajectory using a Markov state model. The region of interest, $[-1, 1]\times [-1, 1]$, is discretized into 
	50-by-50 bins. In this model, we use a lag time of $\tau=100$ for both theoretical analysis and model learning. The transition probability $p_{ij}$ of moving from  bin $i$ to bin $j$ is given by \blauw{Eq.~\ref{transition_prop_ring}}, where $C_i$ is the normalization factor for bin $i$ such that the total transition probability associated with bin $i$ sums to unity.
	
	\begin{equation}\label{transition_prop_ring}
	p_{ij} = \begin{cases}
	C_i e^{-(V_j-V_i) / (k_BT)}, \ \text{if $i$,$j$ are neighbors or $i$=$j$}\\
	0, \ \text{otherwise}.
	\end{cases}
	\end{equation}
	
\begin{figure*}[ht!]
		\begin{center}
			\includegraphics[width=.9\textwidth]{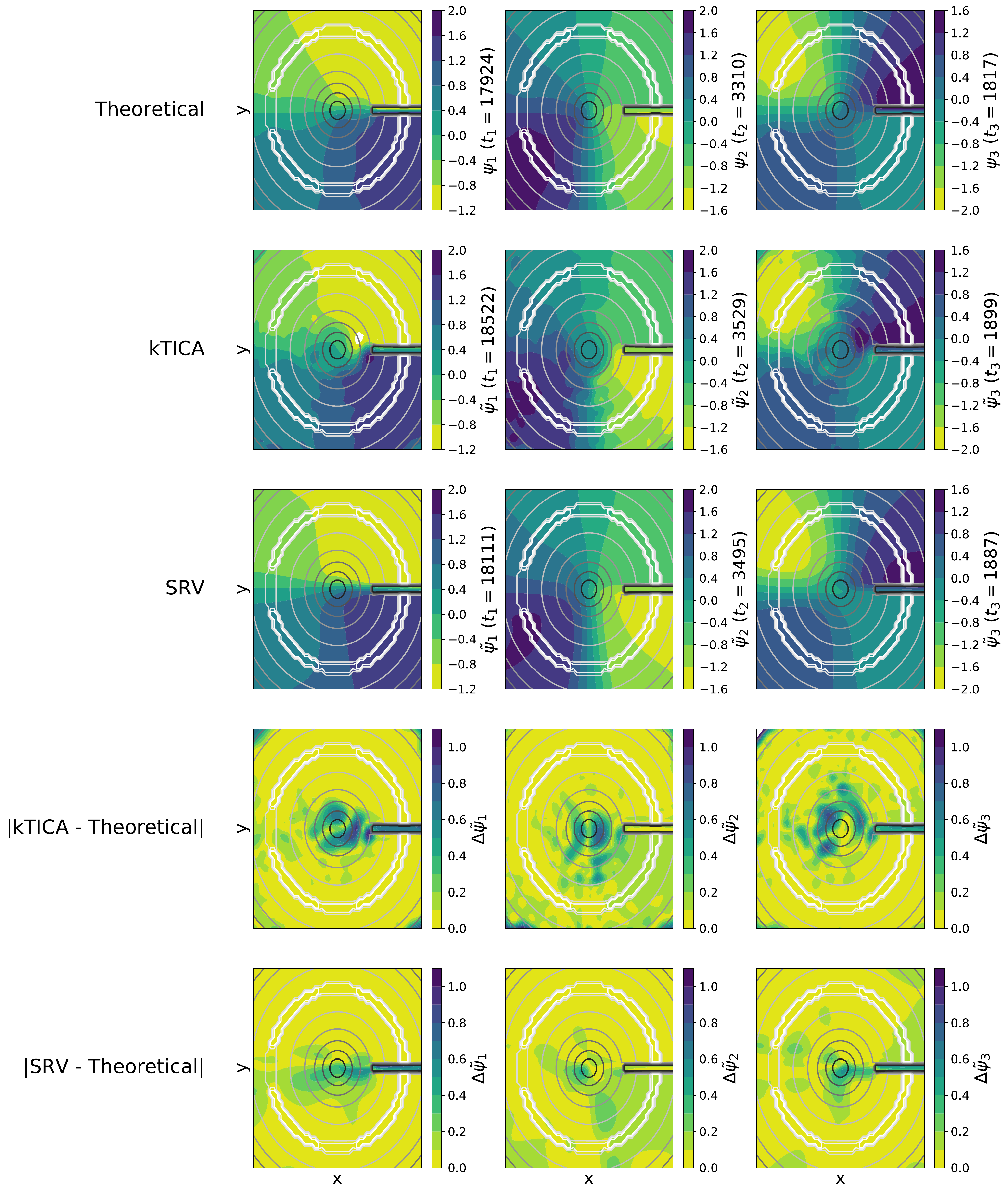}
			\caption{Theoretical eigenfunctions (row 1) and eigenfunctions learned by kernel TICA (row 2) and SRV (row 3) of the 2D ring potential.  Each column denotes one eigenfunction with values represented in colors.  Timescales are shown in corresponding colorbar labels.  Contours of the ring potential are shown in gray to provide a positional reference. Absolute differences with respect to the theoretical eigenfunctions of the kernel TICA (row 4) and SRV (row 5) results.} \label{ring_states}
		\end{center}
	\end{figure*}

	Once again we compute the first three non-stationary theoretical eigenfunctions of $\mathscr{T}(\tau=100)$ from the transition matrix and illustrate these in \blauw{Fig.~\ref{ring_states}}. We then numerically simulate a 5,000,000-step trajectory over the 2D landscape under the action of the transition matrix, and pass these data to a 1000-landmark kTICA model employing a Gaussian kernel and an SRV with the same hidden layer architecture and loss function as the previous example. The state of the system at any time $t$ is defined by the 2D ($x$,$y$)-coordinate pair representing the particle location $(x(t),y(t)) \in \mathbb{R}^2$. Again small deviations between the estimated and theoretical timescales should be expected due to noise and finite length of the simulated data.

\begin{figure}[ht!]
		\begin{center}
			\includegraphics[width=\figurewidth]{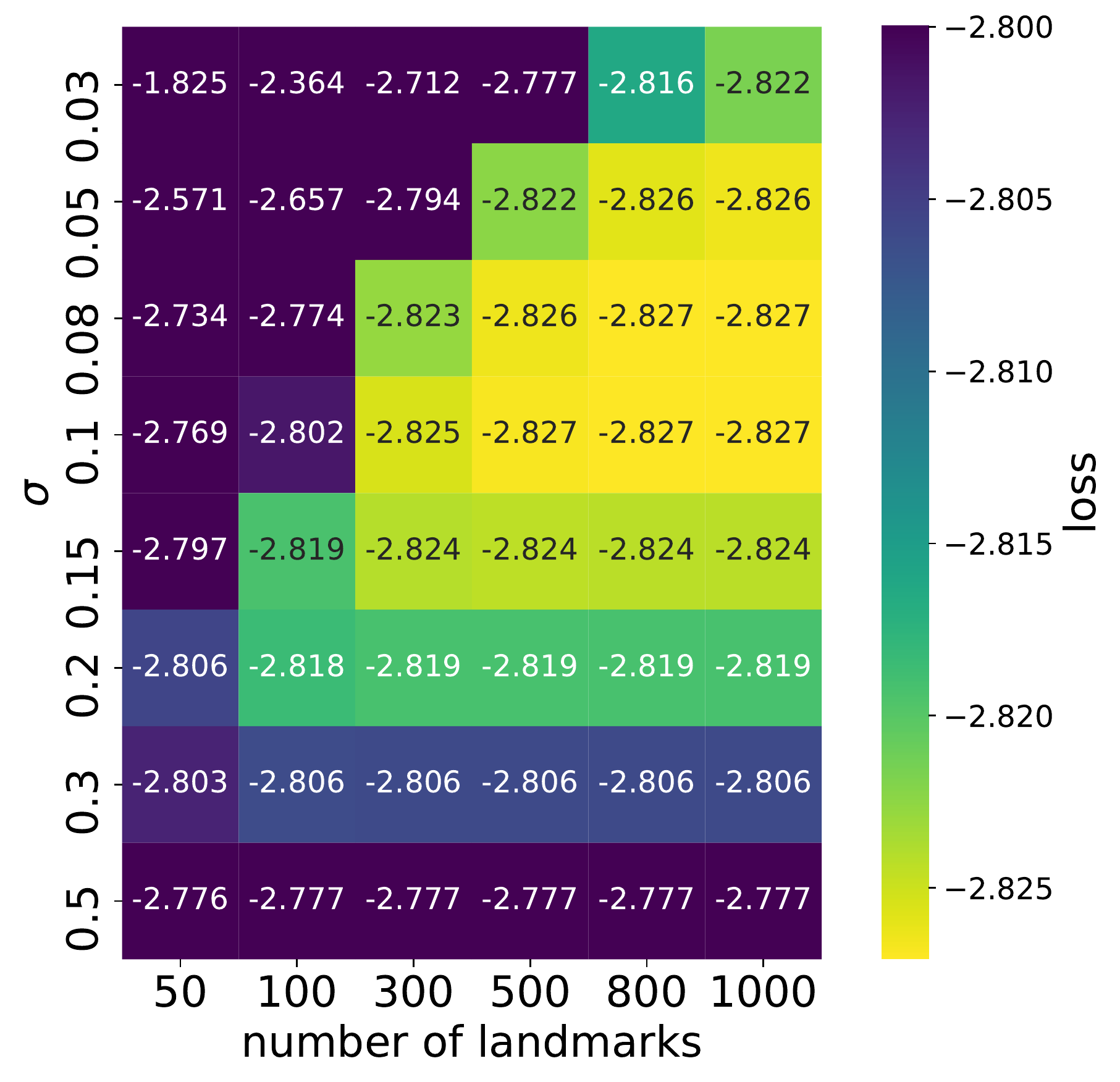}
			\caption{Test loss of kTICA models with different Gaussian kernel bandwidths $\sigma$ and number of landmarks learned by K-means clustering applied to the ring potential. Good test losses (i.e., minimization of the loss function) are only obtained for a large number of landmarks and small $\sigma$.} \label{ring_test_loss}
		\end{center}
	\end{figure}

\begin{figure*}[ht!]
	\begin{center}
		\includegraphics[width=\textwidth]{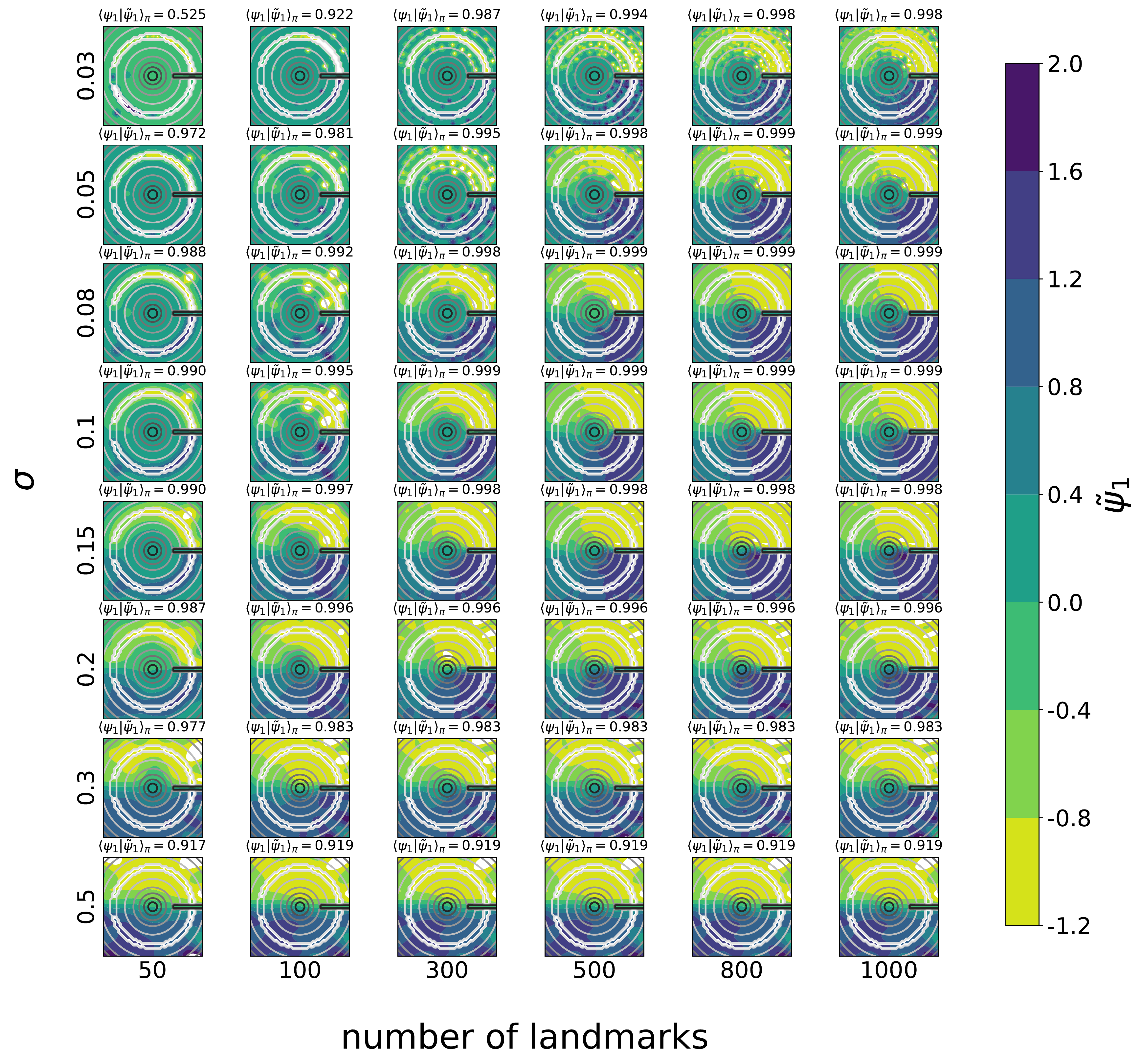}
		\caption{Leading non-trivial eigenfunctions of the 2D ring potential learned by kTICA employing different Gaussian kernel bandwidths $\sigma$ and number of landmarks defined by K-means clustering. The projection coefficient $\langle \psi_1 \vert \tilde{\psi}_1\rangle_\pi$ of the leading learned eigenfunctions on the corresponding leading theoretical eigenfunction are reported directly above each plot as a measure of quality of the kTICA approximation. Contours of the ring potential are shown in gray to provide a positional reference.} \label{ring_test_eigenstates}
	\end{center}
\end{figure*}

	The kTICA results employing the optimal bandwidth $\sigma$ of the Gaussian kernel are shown in \blauw{Fig.~\ref{ring_states}}. 
	Although it gives a reasonable approximation of the eigenfunctions within the ring where data are densely populated, the
	agreement outside of the ring is much worse. This is due to the intrinsic limitation of a Gaussian kernel function: 
	a small $\sigma$ leads to an accurate representation near landmarks but poor predictions for regions far away, while a 
	large $\sigma$ produces better predictions far away at the expense of local accuracy. Moreover, the kTICA results depend sensitively on both the number of landmarks and the bandwidth of the Gaussian kernel.  In \blauw{Fig.~\ref{ring_test_loss}} we report the test loss given by \blauw{Eq.~\ref{loss_f}} on a dynamical trajectory of length 1,000,000 for kTICA models with different kernel bandwidths $\sigma$ and numbers of landmarks selected by K-means clustering. The approximations of the leading non-trivial eigenfunctions are reported in \blauw{Fig.~\ref{ring_test_eigenstates}}.  We note that only when we use a large number of landmarks can we achieve reasonable results, which leads to expensive computations in both landmark selection and calculation of the Gram matrix.  Moreover, even with a large number of landmarks, the range of $\sigma$ values for which satisfactory results are obtained is still quite small and requires substantial tuning.

In contrast, the SRV with architecture of [2, 100, 100, 3] shows excellent agreement with the theoretical eigenfunctions without any tuning of network architecture, activation functions, or loss function \rood{(\blauw{Fig.~\ref{ring_states}})}. The SRV eigenvalues closely match those extracted by kTICA, and the SRV eigenfunctions show superior agreement with the theoretical results compared to kTICA. This result demonstrates the capacity of SRVs in leveraging the flexibility and robustness of neural networks in approximating arbitrary continuous function over compact sets.

	\subsection{Alanine dipeptide}
	
\begin{figure}[ht!]
		\begin{center}
		\includegraphics[width=\figurewidth]{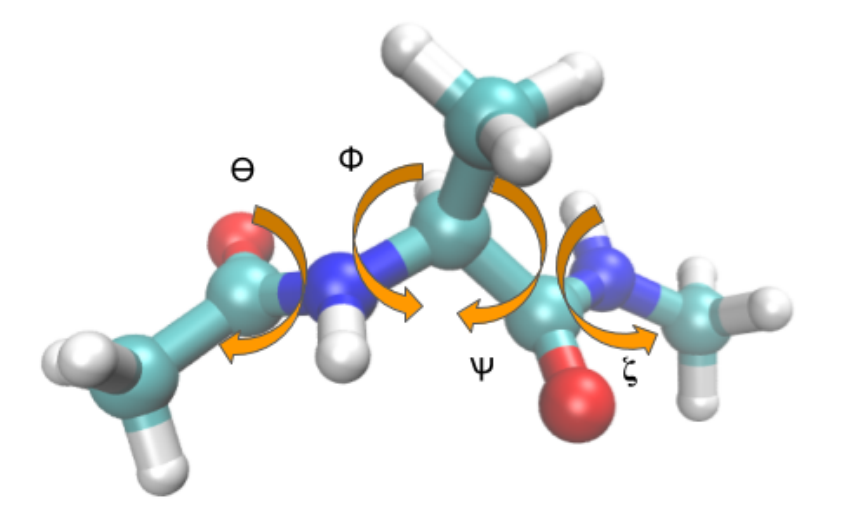}
			\caption{Molecular rendering of alanine dipeptide annotated with the four backbone dihedral angles. Image constructed using VMD~\cite{humphrey1996vmd}.} \label{adp_molecule}
		\end{center}
	\end{figure}

	Having demonstrated SRVs on toy models, we now consider their application to alanine dipeptide in water as a simple but realistic application to molecular data.  Alanine dipeptide (N-acetyl-L-alanine-N$^\prime$-methylamide) is a simple 22-atom peptide that stands as a standard test system for new biomolecular simulation methods. The molecular structure of alanine dipeptide annotated with the four backbone dihedral angles that largely dictate its configurational state is presented in \blauw{Fig.~\ref{adp_molecule}}. 
	A 200 ns  simulation of alanine dipeptide in TIP3P water and modeled using the Amber99sb-ILDN forcefield was conducted at T = 300 K and P = 1 bar using the OpenMM 7.3 simulation suite \cite{eastman2012openmm,eastman2017openmm}.  Lennard-Jones interactions were switched smoothly to zero at
a 1.4 nm cutoff, and electrostatics treated using particle-mesh Ewald with a real space cutoff of 1.4 nm and a reciprocal space grid spacing of 0.12 nm.  Configurations were saved every 2 ps to produce generate a trajectory comprising 1,000,000 configurations. The instantaneous state of the peptide is represented by the Cartesian coordinates of the 22 atoms $x(t) \in \mathbb{R}^{66}$, where the influence of the solvent molecules is treated implicitly through their influence on the peptide configuration. In this case the theoretical eigenfunctions of the transfer operator are unavailable, and we instead compare the SRV results against those of kTICA.

\begin{figure*}[ht!]
		\begin{center}
		\includegraphics[width=\textwidth]{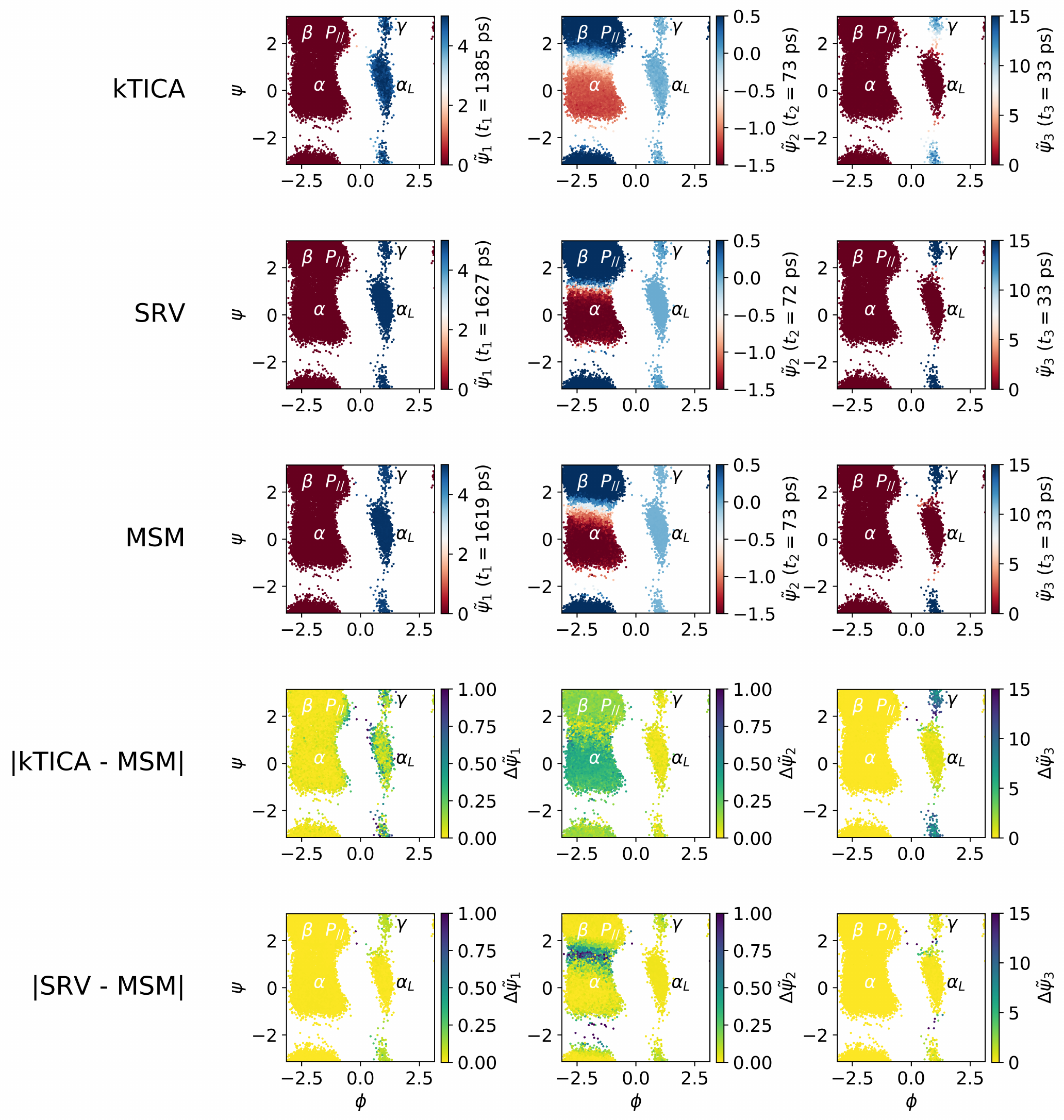}
			\caption{Eigenfunctions of alanine dipeptide learned by kTICA (row 1), SRV (row 2), and MSM constructed over TICA (row 3). Eigenfunctions are superposed as heatmaps over the Ramachandran plot in the $\phi$ and $\psi$ backbone dihedrals. The metastable basins within the plot are labeled according to their conventional terminology.  Timescales of learned eigenfunctions are shown in corresponding colorbar labels. Absolute differences with respect to the MSM eigenfunctions of the kernel TICA (row 4) and SRV (row 5) results.} \label{adp_msm}
		\end{center}
	\end{figure*}

The 45 pairwise distances between the 10 heavy atoms were used as features with which to perform kTICA employing employing a Gaussian kernel, 5000 landmarks obtained from K-means clustering, and a lag time of $\tau$ = 20 ps. The intramolecular pairwise distances or contact matrix are favored within biomolecular simulations as an internal coordinate frame representation that is invariant to translational and rotational motions \cite{sittel2014principal}. The leading three eigenfunctions $\tilde{\psi}_i\ (i=1,2,3)$ discovered by kTICA employing a manually tuned kernel bandwidth are shown in \blauw{Fig.~\ref{adp_msm}} superposed upon the Ramachandran plot in the backbone $\phi$ and $\psi$ torsional angles that are known to be good discriminators of the molecular metastable states \cite{ferguson2011integrating,ren2005transition,laio2002escaping,barducci2008well,bolhuis2000reaction,valsson2014variational,stamati2010application}.  The timescales of the 4$^\mathrm{th}$ and higher order modes lie below the $\tau$ = 20 ps lag time so cannot be resolved by this model. Accordingly, we select three leading slow modes for analysis. From \blauw{Fig.~\ref{adp_msm}} it is apparent that the first slow mode captures transitions along $\phi$ torsion, the second characterizes the transitions between $\alpha$ and $(\beta, P_{//})$ basins, and the third motions between the $\alpha_L$ and $\gamma$ basins.
	
The trajectory is then analyzed at the same lag time using a [45, 100, 100, 3] SRV employing the same hidden layer architecture and loss function as the previous examples. The SRV eigenfunctions illustrated in \blauw{Fig.~\ref{adp_msm}} are in excellent agreement with those learned by kTICA, but importantly the implied timescale of the SRV leading mode is 17\% slower than that extracted by kTICA. What is the origin of this discrepancy?
	
The current state-of-the-art methodology to approximate the eigenfunctions of the transfer operator is to construct a Markov state model (MSM) over a microstate decomposition furnished by TICA \cite{pande2010everything,schwantes2013improvements,pande2010everything,husic2018markov,trendelkamp2015estimation,sultan2017transfer,mittal2018recruiting,harrigan2017msmbuilder,wehmeyer2018introduction,scherer2018variational}  Applying TICA to the 45 pairwise distances between the heavy atoms, we construct a MSM using PyEMMA~\cite{scherer_pyemma_2015}, and present the results in \blauw{Fig.~\ref{adp_msm}}. We see that the MSM eigenfunctions are in excellent accord with those learned by SRVs and kTICA, however while SRVs nearly exactly match the implied timescales of the MSM and results reported in Ref.~\cite{trendelkamp2015estimation}, kTICA substantially underestimates the timescale of the slowest mode.  

The underlying reason for this failure is the that spatial resolution in feature space is limited by number of landmarks, and if the Euclidean distance in feature space does not closely correspond to kinetic distances then it requires much finer spatial resolution to resolve the correct slow modes.  For alanine dipeptide, pairwise distances between heavy atoms are not well correlated with the slowest dynamical modes -- here, rotations around backbone dihedrals -- and therefore landmark kTICA models built on these features have difficulty resolving the slowest mode.  This issue may be alleviated by employing more landmarks, but it quickly becomes computationally intractable on commodity hardware to use significantly more than approximately 5000 landmarks.  Another option is to use features that are better correlated with the slow modes. For alanine dipeptide, it is known that the backbone dihedrals are good features, and if we perform kTICA using these input features we do achieve much better estimates of the implied timescale of the leading mode ($t_1$ = 1602 ps). In general, however, a good feature set is not known \textit{a priori}, and for poor choices it is typically not possible to obtain good performance even for large numbers of landmarks. 

\begin{figure}[ht!]
		\begin{center}
			\includegraphics[width=\figurewidth]{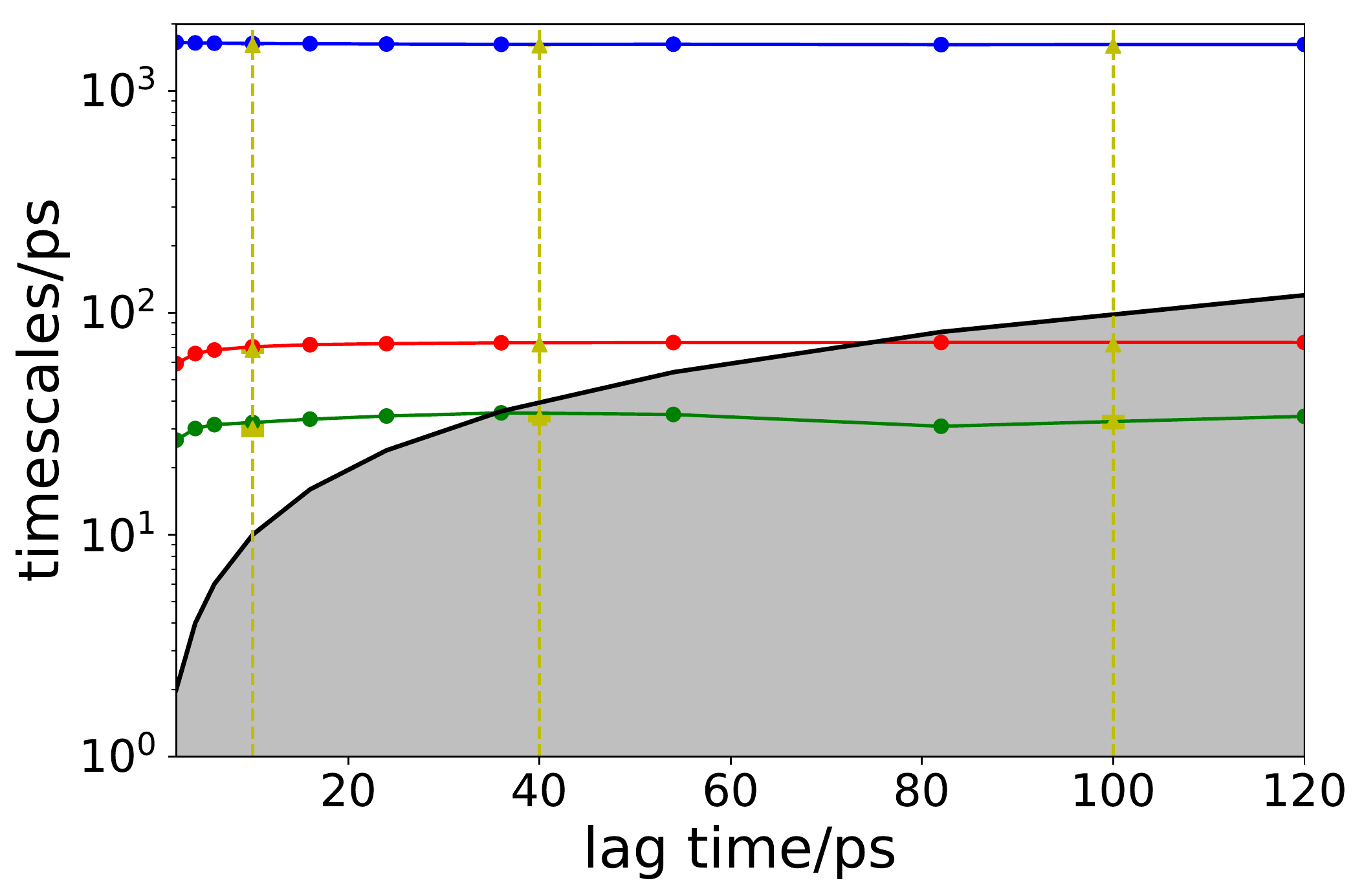}
			\caption{Implied timescales of the three leading eigenfunctions of alanine dipeptide estimated by SRVs employing a variety of lag times. The yellow bars and triangles report the three leading implied timescales for SRVs employing lag times of $\tau$ = 10, 40, and 100 ps. The blue, red, and green traces present the extrapolative estimation of the implied timescales at various lag times using a single SRV constructed with $\tau$ = 20 ps. The solid black line represents the locus of points are which the implied timescale is equal to the lag time. Implied timescales that fall into the shaded area decay more quickly than the lag time and therefore cannot be reliably resolved at that choice of lag time.} \label{adp_other_lag}
		\end{center}
	\end{figure}

Our numerical investigations also show the implied timescales extracted by SRVs to be very robust to the particular choice of lag time. The reliable inference of implied timescales from MSMs requires that they be converged with respect to lag time, and slow convergence presents an impediment to the realization of high-time resolution MSMs. The yellow bars in \blauw{Fig.~\ref{adp_other_lag}} present the implied time scales of the leading three eigenfunctions computed from the mean over five SRVs constructed with lag times of $\tau$ = 10, 40, and 100 ps. It is clear that the implied timescales are robust to the choice of lag time over a relatively large range.

Given the implied timescales evaluated at four different lag times -- $\tau$ = 10, 20, 40, and 100 ps -- this provides an opportunity to assess the dynamical robustness of SRVs by subjecting them to a variant of the Chapman-Kolmogorov test employed in the construction and validation of MSMs~\cite{mardt2018vampnets}. As discussed in \blauw{Section.~\ref{th:transfer_op}}, the VAC framework is founded on the Markovian assumption that the transfer operator is homogeneous in time.  In previous two toy examples, this assumption holds by construction due to the way the trajectory data were generated.  However, for a real system like alanine dipeptide, there is no such guarantee.  Here we test a necessary condition for the Markovian assumption to hold.  In general we have
\begin{equation}\label{key}
\mathscr{T}_t(k\tau)=\prod_{i=0}^{k-1}\mathscr{T}_{t+i\tau}(\tau)
\end{equation}
If the Markovian assumption holds, then the transfer operators are independent of $t$, such that
\begin{equation}\label{ktt_cond}
\mathscr{T}(k\tau)=\mathscr{T}(\tau)^k.
\end{equation}
The corresponding eigenvalue and eigenfunctions are
\begin{align}\label{CK1}
\mathscr{T}(\tau)\circ \tilde{\psi}_{i, \tau}(x)&=\tilde{\lambda}_{i,\tau}\tilde{\psi}_{i, \tau}(x)\nonumber\\
\Rightarrow \mathscr{T}(\tau)^k\circ \tilde{\psi}_{i, \tau}(x)&=\tilde{\lambda}_{i,\tau}^k \tilde{\psi}_{i, \tau}(x),
\end{align}
	and
	\begin{align}\label{CK2}
	\mathscr{T}(k\tau)\circ \tilde{\psi}_{i, k\tau}(x)&=\tilde{\lambda}_{i,k\tau}\tilde{\psi}_{i, k\tau}(x),
	\end{align}
where $\{\tilde{\lambda}_{i,\tau}\}$ and $\{\tilde{\psi}_{i, \tau}(x)\}$ are the estimated eigenvalues and eigenfunctions of $\mathscr{T}(\tau)$, and $\{\tilde{\lambda}_{i,k\tau}\}$ and $\{\tilde{\psi}_{i, k\tau}(x)\}$ those of $\mathscr{T}(k\tau)$. Appealing to \blauw{Eq.~\ref{ktt_cond}}, it follows that
\begin{align}\label{key}
	\tilde{\psi}_{i, k\tau}(x)&=\tilde{\psi}_{i, \tau}(x),\nonumber\\
	\tilde{\lambda}_{i,k\tau}&=\tilde{\lambda}_{i,\tau}^k,
\end{align}
providing a means to compare the consistency of SRVs constructed at different choices of $\tau$. In particular, the implied timescales for the eigenfunctions of $\mathscr{T}(k\tau)$ estimated from an SRV constructed at a lag time $k\tau$
	\begin{equation}\label{key}
		\tilde{t}_{i, \mathscr{T}(k\tau), k\tau}=-\frac{k\tau}{\log \tilde{\lambda}_{i,k\tau}}.
	\end{equation}
should be well approximated by those estimated from an SRV constructed at a lag time $\tau$
\begin{equation}\label{key}
\tilde{t}_{i, \mathscr{T}(\tau), k\tau}=-\frac{k\tau}{\log \tilde{\lambda}_{i,\tau}^k}.
\end{equation}
If this is not the case, then the assumption of Markovianity is invalidated for this choice of lag time.

We present in \blauw{Fig.~\ref{adp_other_lag}} the predicted implied timescales over the range of lag times $\tau$ = 2--120 ps calculated from an SRV constructed at a lag time of $\tau$ = 20 ps. These predictions are in excellent accord to the implied timescales directly computed from SRVs constructed at lag times of $\tau$ = 10, 40, and 100 ps, demonstrating satisfaction of the Chapman-Kolmogorov test and a demonstration of the Markovian property of the system at lag times $\tau \gtrsim 10$ ps \cite{mardt2018vampnets,wehmeyer2018introduction,chodera2006long,pande2010everything,husic2018markov}.

\subsection{WW domain}
	
\begin{figure*}[ht!]
		\begin{center}
			\includegraphics[width=\textwidth]{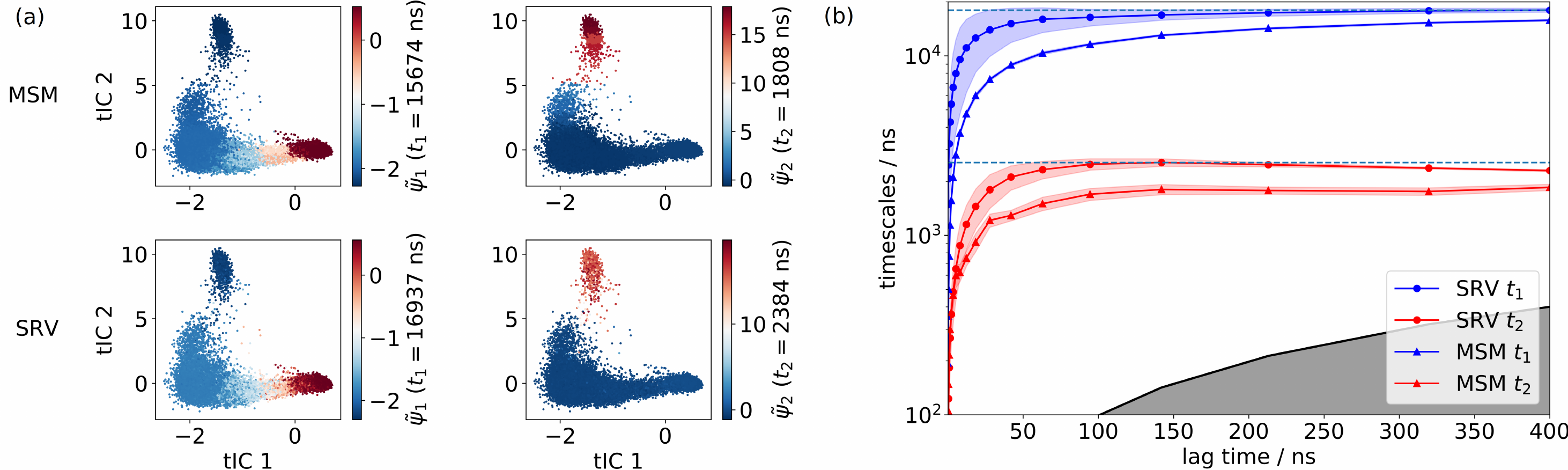}
			\caption{Application of a TICA-based MSM and SRV to WW domain. (a) Eigenfunctions of WW domain learned by TICA-based MSM (row 1) and SRV (row 2) projected onto the two leading TICA eigenfunctions (tIC$_1$, tIC$_2$).  Implied timescales of the learned eigenfunctions are noted adjacent to each plot. (b) Convergence of the implied timescales as a function of lag time for the two leading eigenfunctions of the system as computed by TICA-based MSM (triangles) and SRV (circles). Colored lines indicate the mean implied timescale averages over 10 independently trained models runs and colored shading the 95\% confidence intervals. The teal horizontal dashed lines indicate the converged implied timescales achieved by both the MSM and SRV at large $\tau$.} \label{ww_states}
		\end{center}
	\end{figure*}

	Our final example considers a 1137 $\mu s$ simulation of the folding dynamics of the 35-residue WW domain protein performed in Ref.~\cite{lindorff2011fast}. We use 595 pairwise distances of all C$_\alpha$ atoms to train a TICA-based MSM and a [595, 100, 100, 2] SRV with the same hidden layer architecture and loss function as all previous examples. We use lag time of $\tau$ = 400 ns (2000 steps) for both models, and focus on the two leading slowest modes. The implied timescales of higher-order modes lie close to or below the lag time and so cannot be reliably resolved by this model. The slow modes discovered by the TICA-based MSM and the SRV are shown in \blauw{Fig.~\ref{ww_states}a} projected onto the two leading TICA eigenfunctions (tIC$_1$, tIC$_2$) to furnish a consistent basis for comparison. The MSM and SRV eigenfunctions are in excellent agreement, exhibiting Pearson correlation coefficients of $\rho$ = 0.99, 0.98 for the top two modes respectively. The implied timescales inferred by the two methods are also in good agreement.

	An important aspect of model quality is the convergence rate of the implied timescales with respect to the lag time $\tau$. The lag time must be selected to be sufficiently large such that the state decomposition is Markovian whereby dynamical mixing with states is faster than the lag time and interconversion between states is slower, but it is desirous that the lag time be as short as possible to produce a model with high time resolution \cite{pande2010everything}. Better approximations for the leading eigenfunctions of the transfer operator typically lead to convergence of the implied timescales at shorter lag times. We construct 10 independent SRV models and 10 independent TICA-based MSMs over the WW domain trajectory data and report in \blauw{Fig.~\ref{ww_states}b} the mean and 95\% confidence intervals of the implied timescale convergence with lag time. The SRV exhibits substantially faster convergence than the TICA-based MSM, particularly in the second eigenfunction. This suggests that the eigenfunctions identified by the SRV, although strongly linearly correlated with those identified by the TICA-based MSM, provide a better approximation to the leading slow modes of the transfer operator and produce a superior state decomposition. We attribute this observation to the ability of the SRV to capture complex nonlinear relationships in the data within a continuous state representation, whereas the MSM is dependent upon the TICA coordinates which are founded on a linear variational approximation to the transfer operator eigenfunctions that subsequently inform the construction of an inherently discretized microstate transition matrix from which we compute the MSM eigenvectors. This result demonstrates the viability of SRVs to furnish a high-resolution model of the slow system dynamics without the need to perform any system discretization and at a higher time resolution (i.e., lower lag time) than is possible with the current state-of-the-art TICA-based MSM protocol.
	
	\section{Conclusions} \label{sec:concl}
	
	In this work we proposed a new framework that we term \rood{state-free reversible VAMPnets} (SRV) for the discovery of a hierarchy of nonlinear slow modes
	of a dynamical system from trajectory data. The framework is built on top of transfer operator theory that uses
	a flexible neural network to learn an optimal nonlinear basis from the input representation of the system. Compared to kernel TICA and variational dynamics encoders, our SRV framework has many
	advantages. It is capable of simultaneously learning an arbitrary number of eigenfunctions while guaranteeing orthogonality. 
	It also requires $O(N)$ memory and $O(N)$ computation time, which makes it amenable to large data sets such as those commonly
	encountered in biomlecular simulations. The neural network architecture does not require the selection of a kernel function
	or adjustment of hyperparameters that can strongly affect the quality of the results and be tedious and challenging to tune~\cite{sultan2018transferable, hernandez2018variational}.
	In fact, we find that training such a simple fully-connected feed-forward neural network is simple, cheap, and insensitive to batch 
	size, learning rate, and architecture. Finally, the SRV is a parametric model, which provides an explicit and differentiable mapping from configuration $x$ to
	the learned approximations of the leading eigenvectors of the transfer operator $\{\tilde{\psi}_i\}$. \rood{These slow collective variables are then ideally suited to be utilized in collective variable-based enhanced
	sampling methods where the differentiability of the SRV collective variables enable their seamless incorporation with powerful biased sampling techniques such as metadynamics~\cite{laio2002escaping}.}
	
	The SRV framework possesses a close connection with a number of existing methodologies. In the one-dimensional limit, SRVs are formally equivalent to
	variational dynamics encoders (VDEs) with an exclusive autocorrelation loss, subject to Gaussian noise \cite{hernandez2018variational}. VDEs however, cannot currently generalize to 
	multiple dimensions due to the lack of an orthogonality constraint on the learned eigenfunctions. By using the more general 
	VAMP principle for non-reversible processes and targeting membership state probabilities rather than learning continuous 
	functions, VAMPnets are obtained \cite{mardt2018vampnets}.

In regards to the analysis of molecular simulation trajectories, we anticipate that the flexibility and high time-resolution of SRV models will be of use in helping resolve and understand the important long-time conformational changes governing biomolecular folding and function. Moreover, it is straightforward to replace TICA-based MSMs with SRV-based MSMs, maintaining the large body of theoretical and practical understanding of MSM construction while delivering the advantages of SRVs in improved approximation of the slow modes and superior microstate decomposition. In regards to enhanced sampling in molecular simulation, the differentiable nature of SRV coordinates naturally enables biasing along the SRV collective variables (CVs) using well-established accelerated sampling techniques such as umbrella sampling, metadynamics, and adaptive biasing force. The efficiency of these techniques depends crucially on the choice of ``good'' CVs coincident with the important underlying dynamical modes governing the ling-time evolution of the system. A number of recent works have employed neural networks to learn nonlinear CVs describing the directions of highest variance within the data  \cite{chen2017mesa,chen2018collective,ribeiro2018reweighted,ribeiro2018towards}. However, the high variance directions are not guaranteed to also correspond to the slow directions of the dynamics. Only variational dynamics encoders have been used to learn and bias sampling in a slow CV \cite{sultan2018transferable}, but, as observed above, the VDE is limited to approximate only the leading eigenfunction of the transfer operator. SRVs open the door to performing accelerated sampling within the full spectrum of all relevant eigenfunctions of the transfer operator. In a similar vein, SRVs may also be profitably incorporated into adaptive sampling approaches that do not apply artificial biasing force, but rather smartly initialize short unbiased simulations on the edge of the explored domain \cite{shamsi2018reinforcement, preto2014fast, weber2011characterization, zimmerman2015fast,bowman2010enhanced,hinrichs2007calculation,doerr2014fly}. The dynamically meaningful projections of the data into the SRV collective variables is anticipated to better resolve the dynamical frontier than dimensionality reduction approaches based on maximal preservation of variance, and therefore better direct sampling along the slow conformational pathways. In sum, we expect SRVs to have a variety of applications not just in the context of molecular simulations, but also more broadly within the analysis of dynamical systems. 


	\section*{Acknowledgments}
	
	This material is based upon work supported by the National Science Foundation under Grant No.~CHE-1841805.
	H.S. acknowledges support from the Molecular Software Sciences Institute (MolSSI) Software Fellows program 
	(NSF grant ACI-1547580) \cite{krylov2018perspective,wilkins2018nsf}. We are grateful to D.E. Shaw Research for sharing the WW domain simulation trajectories.

	\section*{Software availability}
	
	Software to perform slow modes discovery along with an API compatible with scikit-learn and Keras, Jupyter notebooks to reproduce the results presented in this paper, documentation, and examples have been freely available under MIT license at \url{https://github.com/hsidky/srv}.

	\clearpage
	\newpage
	
	\onecolumngrid

	\bibliography{library}

	

\end{document}